%% file: article.tex
\pdfoutput=1

\documentclass[11pt]{article}

\usepackage{acl}
\usepackage{times}
\usepackage{latexsym}

\usepackage[T1]{fontenc}
\usepackage[utf8]{inputenc}

\usepackage{multirow}

\usepackage{microtype}

\usepackage{tikz,pgfplots,pgfplotstable}
\usepackage{booktabs}
\usepackage{colorblind}
\usepackage{soul}
\DeclareRobustCommand{\hlc}[2]{{\sethlcolor{#1}\hl{#2}}}

\usepackage{inconsolata}

\usetikzlibrary{shapes.misc, positioning, decorations.pathreplacing, calc}
\usepackage{subcaption}

\usepackage{tcolorbox}
\definecolor{azure}{rgb}{0.0, 0.5, 1.0}
\tcbuselibrary{skins}
\tcbset{enhanced}
\newcommand{\qsecbox}[1]{\begin{tcolorbox}[left=1mm,right=1mm,boxrule=0.2mm,leftrule=2mm,drop fuzzy shadow,colframe=lightgray,frame style={left color=azure!90!lightgray}]#1\end{tcolorbox}}
\newcommand{\egcvalue}[1]{\textbf{\textit{#1}}}

\usepackage{enumitem}
\usepackage{amssymb}

\usepackage{amsmath}

\usepackage{pgf}

\usepackage{float}

\usepackage[font=footnotesize,labelfont=bf]{caption}

\definecolor{tgb1}{rgb}{0.2666, 0.466, 0.666}
\definecolor{tgb2}{rgb}{0.4, 0.8, 0.9333}
\definecolor{tgb3}{rgb}{0.1333, 0.5333, 0.2}
\definecolor{tgb4}{rgb}{0.8, 0.733, 0.2666}
\definecolor{tgb5}{rgb}{0.9333, 0.4, 0.466}
\definecolor{tgb6}{rgb}{0.666, 0.2, 0.466}

\usepackage{hyperref}
\usepackage[english]{babel}
\addto\extrasenglish{%

}

\title{\texttt{JUDGEBERT}: Assessing Legal Meaning Preservation Between Sentences}

\author{David Beauchemin\textsuperscript{\textdagger}, Michelle Albert-Rochette\textsuperscript{\textdaggerdbl}, Richard Khoury\textsuperscript{\textdagger} \and Pierre-Luc D\'eziel\textsuperscript{\textdaggerdbl}\\
Universit\'e Laval, Qu\'ebec, Canada \\
Computer Science Department\textsuperscript{\textdagger} and Faculty of Law\textsuperscript{\textdaggerdbl}\\
\texttt{david.beauchemin@ift.ulaval.ca}, \texttt{michelle.albert-rochette.1@ulaval.ca}\\
\texttt{richard.khoury@ift.ulaval.ca}, \texttt{pierre-luc.deziel@fd.ulaval.ca}
}

\newcommand{\guillemet}[1]{``#1''}

\newcommand{\david}[1]{\textcolor{red}{(#1 - DB)}}

\begin{document}
\maketitle
\begin{abstract}
Simplifying text while preserving its meaning is a complex yet essential task, especially in sensitive domain applications like legal texts. 
When applied to a specialized field, like the legal domain, preservation differs significantly from its role in regular texts.
This paper introduces FrJUDGE, a new dataset to assess legal meaning preservation between two legal texts.
It also introduces \texttt{JUDGEBERT}, a novel evaluation metric designed to assess legal meaning preservation in French legal text simplification. 
\texttt{JUDGEBERT} demonstrates a superior correlation with human judgment compared to existing metrics.
It also passes two crucial sanity checks, while other metrics did not: For two identical sentences, it always returns a score of 100\%; on the other hand, it returns 0\% for two unrelated sentences.
Our findings highlight its potential to transform legal NLP applications, ensuring accuracy and accessibility for text simplification for legal practitioners and lay users.
\end{abstract}

\section{Introduction}
Automatic text simplification (ATS) aims to  creates easier-to-read text while keeping the original meaning \citep{saggion2017automatic}.
Evaluating whether a simplified text preserves the meaning of the original complex one is not trivial. Yet, it is critical for ATS and many other natural language processing (NLP) tasks, such as machine translation \citep{gatt2018survey}. 
Evaluation of ATS is based on three dimensions of system generations: \guillemet{fluency}, \guillemet{simplicity} and \guillemet{meaning preservation}.
Fluency measures grammatical correctness, simplicity estimates how easy-to-understand the text is, while meaning preservation measures how well the output text's meaning corresponds to the original \citep{saggion2017automatic}.
It is typical to use automatic metrics to assess these evaluations, such as BLEU \citep{papineni2002bleu} and SARI \citep{xu2015problems}, which tend to focus on only one of the three dimensions.
For example, BLEU is commonly used to evaluate fluency, while SARI is for simplicity.
More recent automatic NLP metrics use Transformer architecture to compute the ATS. For example, MeaningBERT \citep{beauchemin2023meaningbert} is an evaluation metric that uses a fine-tuned BERT Transformer model to assess meaning preservation.

When applied to specialized fields, such as the legal domain, preservation differs significantly from its role in regular texts and has the potential to significantly impact all stakeholders.  
Inaccurate ATS can mislead users, cause legal issues, or represent a risk for the company that deploys the system \cite{vstajner2021automatic}.
A user can interpret an automatically-simplified text in a way that would not hold in court, creating a \guillemet{legal gap} between the texts' meanings.
For example, in 2024, an Air Canada passenger was misled about the airline's rules for bereavement fares when the company's AI chatbot hallucinated an answer inconsistent with their policies. 
The Tribunal found Air Canada guilty of \guillemet{negligent misrepresentation} in this situation \cite{air_Canada_court}.
None of the metrics currently available specifically assess legal meaning preservation and have not been benchmarked against human judgment for this sensitive task. 
Developing a metric to assess whether a simplification still conveys the same legal meaning is crucial to minimizing risk in legal NLP applications.
To this end, we introduced \guillemet{legal meaning} as a substitute to \guillemet{meaning preservation} to evaluate ATS system output for legal text simplification (TS). Our three contributions are:

\begin{enumerate}[leftmargin=*, noitemsep, topsep=0ex]
    \item We proposed a new dimension to evaluate ATS system output for legal ATS;
    \item We proposed \href{https://github.com/GRAAL-Research/JUDGEBERT}{FrJUDGE}\footnote{\href{https://github.com/GRAAL-Research/JUDGEBERT}{https://github.com/GRAAL-Research/JUDGEBERT}}, a \textbf{Fr}ench corpus of insurance legal meaning \textbf{JUDGE}ments to assess legal meaning preservation between an insurance contract and its simplified text; and
    \item \texttt{JUDGEBERT}, a new fine-tuned BERT metric designed to assess the legal meaning preservation between two French legal sentences, which we have trained to correlate with human judgment on French insurance text. 
\end{enumerate}

This paper is outlined as follows: first, we study the relevant ATS metrics research and corpora in \autoref{sec:related}. 
Then, we propose a definition of legal meaning, and we present our corpus in \autoref{sec:judge}, along with our new trainable metric, \texttt{JUDGEBERT} in \autoref{sec:judgebert}. To demonstrate its quality, we also present a set of experiments in \autoref{sec:exp}, and, following \citet{beauchemin2023meaningbert}, we will also conduct a set of sanity checks. 
Finally, we will discuss our results in \autoref{sec:metrics_analysis} and conclude in \autoref{sec:conclusion}.

\section{Related Work}
\label{sec:related}

\subsection{Human Evaluation and Automatic Metrics for Meaning Preservation}
Since automatic metrics are a proxy for human judgments for ATS, they should correlate well with human ratings. 
However, \citet{sulem2018bleu} found low to no correlation between BLEU and meaning preservation dimensions when sentence splitting is involved, a typical simplification operation used notably for legal texts \cite{garimella2022text}.
They also pointed out that BLEU is sensitive to the length of the compared texts and does not consider semantic variability between sentences that differ on synonymous words or in word order.

Since word-embeddings-based metrics can better account for semantic variability between sentences \citep{zhang2019bertscore}, \citet{beauchemin2023meaningbert} have conducted a correlation analysis on meaning preservation of 22 ATS metrics.
These include popular non-Transformer and Transformer ones.
Their results show that many non-Transformer metrics correlate poorly with human judgment, and most Transformer ones correlate weakly. 
In addition, they also conducted benchmarking tests to evaluate meaning preservation between pairs of identical and unrelated sentences. 
These tests show that many automatics metrics fail even in these simple tasks.
Furthermore, they proposed MeaningBERT, a fine-tuned Transformer-based metric that correlated better with human judgment and passed the benchmarking tests.
Nevertheless, none of these metrics focus on legal meaning.

\subsection{French Legal Text Simplification Datasets}
\label{subsec:frdatasets}
Only three TS French datasets are available in the literature, and none focus on legal documents \cite{ryan2023revisiting}.
Indeed, Alector \cite{gala2020alector} focuses on literacy and scientific texts, while CLEAR \cite{grabar2018clear} on medical text, and WikiLarge FR \cite{cardon2020french} on informative text from Wikipedia.
None of these available corpora are suited to our needs since legal documents differ from other texts: they are lengthier and use specialized vocabulary \cite{katz2023natural}.
Only two corpora of legal documents are available in French: RISCBAC \cite{Beauchemin2023RISC}, a set of synthetic bilingual automobile insurance legal contracts, and EUR-Lex-Sum \cite{aumiller2022eur}, a multi- and cross-lingual set of summaries of legal acts from the European Union law platform. However, neither dataset includes simplifications or human annotations.

\section{FrJUDGE: a French Corpus of Insurance Legal Meaning Judgments}
\label{sec:judge}
In this section, we introduce the \textbf{Fr}ench corpus of insurance legal meaning \textbf{JUDG}m\textbf{E}nts (FrJUDGE), which is the first legal meaning judgment dataset in any language.
FrJUDGE consists of 297 human-annotated French sentences taken from property damage insurance forms used by two insurance regulators, namely the \textit{Bureau d'assurance du Canada} \cite{bac}, and the \textit{Autorité des marchés financiers du Québec} \cite{amf}.
As illustrated in \autoref{tab:frjudge}, each dataset instance consists of a legal sentence, a simplification, and human annotations (simplicity, characterization and legal meaning).
Both sources are publicly available online, and we obtained authorization to publish them under a CC-BY 4.0 license.

\begin{table*}
    \centering
    \resizebox{\textwidth}{!}{%
    \begin{tabular}{p{0.45\textwidth}p{0.45\textwidth}ccc}
    %\toprule
    Legal Sentence & Simplified Sentence  & \texttt{Simplicity Level} & \texttt{Characterization}& \begin{tabular}[c]{@{}c@{}}\texttt{Legal}  \texttt{Meaning}\end{tabular}   \\\midrule
    \textit{L’assuré désigné est le propriétaire réel et le titulaire de l’immatriculation du véhicule désigné.} & \textit{L'assuré désigné possède le véhicule désigné. Il détient aussi son immatriculation.} & \textit{Aussi simple à lire}& 2 & 8 \\
    \end{tabular}%
    }
    \caption{Example of an instance from FrJUDGE containing a legal sentence and human annotations (simplification, simplicity level, characterization and legal meaning).}
    \label{tab:frjudge}
\end{table*}

\subsection{Legal Meaning}
We argue that \guillemet{meaning} and \guillemet{legal meaning} differ because, for typical ATS, synonyms can be used to convey the same meaning, while for legalese, synonyms do not necessarily convey the same meaning.
For example, in the common language, \guillemet{automobile} and \guillemet{vehicle} share the same meaning. However, for legalese, the first means any vehicle moved by a \guillemet{mechanical force} and the latter by \guillemet{mechanical or human force} \cite{loicoderoute}. Thus, an automobile is a vehicle, but a vehicle is not necessarily an automobile. For example, a bicycle is a vehicle that fits the description of a vehicle but not an automobile.

Given that no previous work nor automatic metric focuses on legal meaning and that our goal is to determine whether or not ATS systems can simplify texts while maintaining their meaning from a legal standpoint, we must rigorously define what \guillemet{legal meaning} is.
Only a few articles focus on legal TS, and none specifically study the preservation of legal meaning between two texts. 
However, \citet{hagan2023good} proposes 22 actionable criteria for legal question-answering that any legal AI system should be benchmarked on to fully assess its capabilities and limitations, and guide policymakers and regulators.
These criteria are closely related to \guillemet{how a professional lawyer should conduct themselves in their practice}.
Two of these criteria are particularly interesting for our work: a \guillemet{response is robust and comprehensive, covering details and exceptions} and a \guillemet{response does not misrepresent the substantive law}. Using these two criteria, we proposed the following definition for \guillemet{legal meaning} as a metric to assess the quality of a legal ATS system: \guillemet{\textbf{Legal meaning measures how well the output text conveys the legal details and exceptions and does not misrepresent the law}}.

\subsection{FrJUDGE Corpus}
\subsubsection{Data Collection}
Sentences in FrJUDGE were collected manually from the two insurance forms.
Specifically, we examined all sentences and extracted 312 text blocs based on three criteria: text blocs are

\begin{enumerate}[leftmargin=*, noitemsep, topsep=0ex]
    \item between 1 and 5 sentences long;
    \item not boilerplate texts such as a title; and
    \item college-level reading level grade ($\leq 50$) on the French Flesch-Kincaid grade level (FKGL) \cite{kandel1958application} to focus on more challenging sentences in an insurance contrat.
\end{enumerate}

For example, the sentence (translated) \guillemet{The city and province of the address written in this section 1 constitute the designated vehicle's principal place of use, storage and parking.} passes the first two criteria. However, it scored 69.87 on the FKGL, so it was not selected.

\subsubsection{Automatic Text Simplification}
\label{subsec:ats}
Since few ATS systems exist in French and none are designed explicitly for legal texts, no pre-trained models are available to generate French ATS.
Thus, all sentences in the corpus were automatically simplified using the OpenAI GPTs model through their API.
We selected this approach since it has been shown by \citet{feng2023sentence, kew2023bless, wu2024depth} that foundational large language models (LLMs) generate less erroneous simplification outputs than state-of-the-art approaches; thus, they are effective ATS systems, even when using zero-shot prompting.
Moreover, it also has been shown by \citet{nozza2023really, madina2024preliminary} that GPTs are effective ATS systems in languages other than English, such as Italian and Spanish.
We present the details used for generation in \autoref{ann:prompt} and examples in \autoref{ann:generation}.

\subsubsection{Human Evaluation Methodology}
\label{subsubsec:automatic_eval}
Following the arguments of \citet{van-der-lee-etal-2019-best}, we present our human evaluation methodology's in this section.

\paragraph{Selected models.} We selected GPT4-turbo with zero-shot prompting.

\paragraph{Number of outputs.} 
We randomly selected 297 instances for annotation and 15 for practice\footnote{These practice annotations have been used to help annotator practice their tasks and adjust our guidelines; these examples have been discarded from the final dataset.}.

\paragraph{Presentation and interface.} 
We used a customized version of the Prodigy annotation tool \cite{prodigy_montani_honnibal}, and we present in \autoref{ann:clf_annotation} the interface (in French).
Annotators use our annotation procedure to annotate each instance randomly. Like the ATS system, annotators were not given the overall legal documents. 

\paragraph{Annotators.}
We selected five native French-speaking law students at the Faculty of Law of University Laval as our annotators.
A meeting was held with them to introduce the task, instructions, and annotation guide and interface.
Instructions included that they must spend at most 5 minutes per sentence pair.
Furthermore, 15 instances were annotated during a pilot phase to familiarize them with the task.
Finally, during a second meeting after evaluating the practice instance, annotators received feedback and advice on what phenomena they should be cautious about.
Recognizing the significant contribution of our annotators, they were remunerated fairly according to the University's hourly salary pay scale.
Each annotator completed their work in at most 30 hours.
We provide in-depth details of the evaluation setup in our Human Evaluation Datasheet \cite{shimorina2021human} in \autoref{ann:hed}.

\paragraph{Legal Meaning Scale.}
Given that we have previously defined what the term \guillemet{legal meaning} means, we now define \guillemet{\textbf{legal meaning metric}} as the metric that measures the legal meaning between the legal original and the simplified text. We will refer to this metric as the \texttt{legal meaning preservation} (\texttt{LMP}).
To this end, we designed a Likert scale \cite{Likert1932ATF} ranging from one to ten. 
A simplified text that scores a ten means that the legal meaning between the two texts \guillemet{tends to be preserved}\footnote{Since most of our annotators were reluctant to state that the two sentences were equivalent, and due to the legal risk of stating that a sentence is \guillemet{perfectly preserving its legal meaning}, we choose to be less assertive in our scale.}.
On the other hand, one receiving a score of 1 does not match the original legal meaning at all.

\paragraph{Annotation Procedure.}
The annotators must determine one of TS's three dimensions, simplicity, along with our new fourth dimension to replace the \guillemet{meaning} dimension.
We choose not to evaluate fluency since \citet{wu2024depth} have shown that GPT4 fluency capabilities are near perfect (2.98/3).

First, the annotators assess the \texttt{simplicity} of the simplified text.
We have adopted a simpler version of the eight-level ordinal scale proposed by \citet{Primpied2022Quantifying}, which uses intuitive perception levels of text difficulty ranging from children's stories (lowest) to legal documents (highest). 
Our initial pilot found the scale to be too complex for our case, which is coherent with \citet{stodden2021scale} conclusion that \guillemet{interpretation of the simplicity scale is consistent when rated by experts [...]}.
Indeed, we found that our annotators tended to assign mostly either the \guillemet{legal documents} level or a level below this.
Moreover, annotators expressed their concern about the scale, which motivated us to change it.

Thus, our version uses four levels (translated): \guillemet{Easier to read}, \guillemet{Equal to read}, \guillemet{More difficult}, and \guillemet{No simplification}\footnote{\guillemet{No simplification} applied to the case where the simplification is identical to the original text or in another language.}. 
Since our annotators are legal experts, we selected this approach because expert annotators tend to \guillemet{inject their own opinions and biases} during annotations \cite{van-der-lee-etal-2019-best}. 

Second, following the work of \citet{garneau-etal-2022-evaluating}, the annotators use a three-step process to assess an instance's \texttt{LMP}. First, they decide the \texttt{characterization} of the text.  
Characterization refers to qualifying laws \cite{frechette2010qualification}\footnote{It is worth mentioning that since the sentence is isolated, it can be challenging to select the proper characterization and sometimes more than one can apply. For the latter, annotators must select the one that seems to apply the most.}. 
For example, risks that are not covered in an insurance contract, such as nuclear damage, are characterized as \guillemet{exclusions or restrictions}.
Each annotated sentence is characterized into one of our 18 classes detailed in \autoref{ann:charac}.
This step helps annotators identify the type of legal text the instance refers to; it does not impact the \texttt{LMP} score. 
With this approach, our annotators can rely on their legal background and education to assess whether a simplification respects a class's characterization elements, such as whether it states a proper definition that respects Quebec legislation. 
In the second step, the annotators assign a preliminary \texttt{LMP} score. \citet{garneau-etal-2022-evaluating} observed that legal experts naturally divide their decisions into three regions instead of directly assessing a score between 1 and 10. Consequently, following their work, we split our legal accuracy scale into the three score brackets listed below.

\paragraph{\colorbox{ForestGreen!30}{7}-\colorbox{ForestGreen!80}{10} -- Accurate.} Means the simplification seems to entail the legal details and exceptions properly and does not misrepresent the law; it is considered to \guillemet{tends to be} accurate.
\paragraph{\colorbox{yellow!30}{2}-\colorbox{yellow!60}{6} -- Seems Imprecise.} Means the generation seems to improperly entail the legal details, exceptions and slightly misrepresents the law.

\paragraph{\colorbox{red!50}{1} -- Off-Track.} Means the simplification is obviously erroneous, does not entail the legal details and exceptions, and misrepresents the law.

Once the annotators have chosen the score bracket where the simplification belongs, they move on to the final step: looking for legal errors in the output. We identify four types:

\begin{itemize}[leftmargin=*, noitemsep, topsep=0ex]
    \item \textbf{Hallucinations} are facts the model generates despite not appearing in the original text.
    For example, the simplified text might specify that the insured is covered for a particular risk, while the original clause does not mention it.
    \item \textbf{Omissions} occur when essential facts are in the original text but are not in the simplified text generated by the model.
    For example, the original clause might specify a maximum coverage amount, but the simplified text does not.
    \item \textbf{Consistency} issues occur when the model simplifies a juridical term but does not use the simplified term for all occurrences of the juridical term. For example, if the original clause refers to an \guillemet{automobile} and the simplification replaces it with \guillemet{vehicle}, we do not consider it a consistency error.
    On the other hand, if the simplification alternates between using \guillemet{automobile} and \guillemet{vehicle}, it would be an error.
    \item \textbf{Confusions}: factual mistakes characterized by mismatches between the source and the generation. For example, the source says that the insured must declare claims as soon as possible, but the generation states otherwise.
\end{itemize}

\begin{figure*}
    \centering
    \begin{tikzpicture}
        \fill[fill=red, opacity=0.4] (0, 0) rectangle (1.5, 0.25);
        \fill[fill=yellow, opacity=0.6] (1.5, 0) rectangle (9, 0.25);
        \fill[fill=ForestGreen, opacity=0.4] (9, 0) rectangle (13.5, 0.25);
        \fill[fill=ForestGreen, opacity=0.8] (13.5, 0) rectangle (15, 0.25);
        \draw[draw=black] (0, 0) rectangle (15, 0.25);
        
        \node[below=1mm, align=center] at (0, 0){\scriptsize 1};
        \node[below=1mm, align=center] at (9, 0){\scriptsize 6};
        % \node[below=2mm, align=center] at (8.25, 0){6};
        \node[below=1mm, align=center] at (15, 0){\scriptsize 10};
        
        \node[below=0.4cm, align=center] at (0.25, 0){\scriptsize Off-track};
        \draw[decoration={brace, mirror, amplitude=10pt}, decorate, below=0.35cm, align=center] (1.5, 0) -- node[below=0.5cm] {\scriptsize Various juridical errors} (8.95, 0);
        \draw[decoration={brace, mirror, amplitude=10pt}, decorate, below=0.35cm, align=center] (9.05, 0) -- node[below=0.5cm] {\scriptsize Minor juridical errors} (15, 0);

        \node[below=0.4cm, align=center] at (14.75, 0){\scriptsize Tends to be preserved};
    \end{tikzpicture}
    \caption{The four-section Likert legal meaning scale used for annotation. The annotators first decide where the generation sits between the four regions. Then they remove points for every error encountered.}
    \label{fig:likert_scale}
\end{figure*}
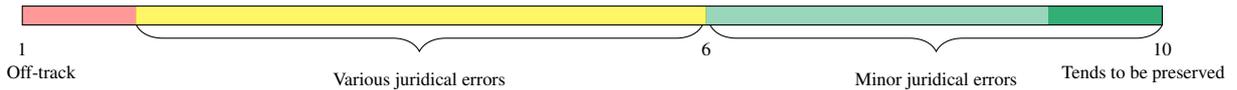

Each error reduces the output's score by one point, starting from the bracket's maximum. The output's score can never drop below the score one (1). To summarize this process, \autoref{fig:likert_scale} conceptualizes the corresponding Likert scale.

\subsubsection{Annotation Results}
\label{subsec:annotation_res}
\begin{figure*}[ht!]
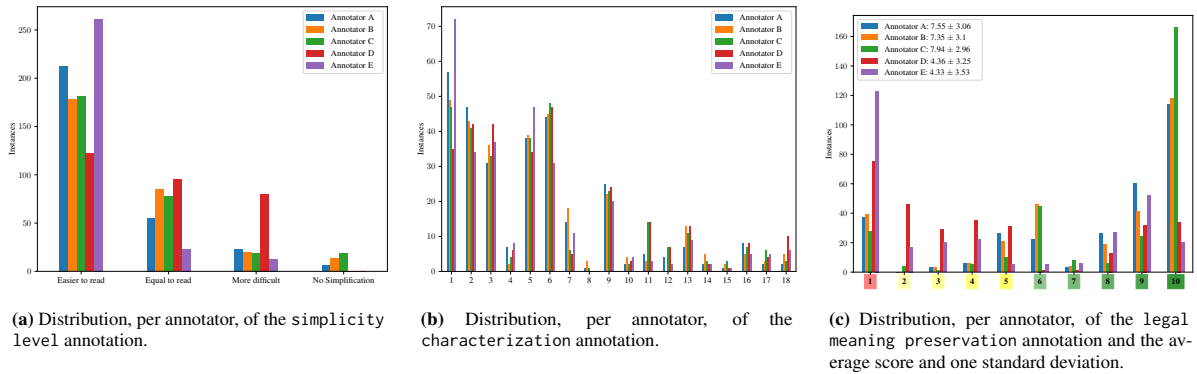

    \centering
    \begin{subfigure}[t]{0.33\textwidth}
        \centering
        \resizebox{\textwidth}{!}{%
            \input{fig/difficulty_level.pgf}
        }
        \captionsetup{width=.925\linewidth}
        \caption{Distribution, per annotator, of the \texttt{simplicity level} annotation.}
        \label{fig:diff_level}
    \end{subfigure}
    \begin{subfigure}[t]{0.33\textwidth}
        \centering
        \resizebox{\textwidth}{!}{%
            \input{fig/qualification.pgf}
        }
        \captionsetup{width=.925\linewidth}
        \caption{Distribution, per annotator, of the \texttt{characterization} annotation.}
        \label{fig:qualification}
    \end{subfigure}
    \begin{subfigure}[t]{0.32\textwidth}
        \centering
        \resizebox{\textwidth}{!}{%
            \input{fig/legal_precision.pgf}
        }
        \captionsetup{width=.925\linewidth}
        \caption{Distribution, per annotator, of the \texttt{legal meaning preservation} annotation and the average score and one standard deviation.}
        \label{fig:legal_precision}
    \end{subfigure}
        \caption{Distribution, per annotator, of the annotation for all three aspects.}
        \label{fig:distribution}
\end{figure*}

We provide in \autoref{fig:distribution} the breakdown, by annotator, of all annotation criteria. 
First, we can see in \autoref{fig:diff_level} and \autoref{fig:qualification} that \texttt{simplicity level} and \texttt{characterization} are distributed similarly among all annotators, except for the annotator E. 
Indeed, annotator E finds more frequently simplified text easier to read than the other annotators and assigns most of its \texttt{characterization} annotations into the first class (i.e. description of endorsement). 
Since this class can act as a generic class, this characterization is adequate but could be more precise. However, since this step acts as an intermediary one, it does not negatively affect the quality of the annotations.
Second, we can also see that for \texttt{LMP}, we seem to have two clusters of similar annotation distributions. 
Annotators A, B and C generally assign similar scores to each other, while annotators D and E often behave similarly to each other but differently from the first group.
Furthermore, the first group has a higher frequency of perfect scores (\colorbox{ForestGreen!80}{10}). On the other hand, annotators D and E more frequently attribute \guillemet{Off-Track} score (\colorbox{red!50}{1}), meaning they were more strict in their initial reading of the simplification. 
This could be due to the annotators' domain expertise, which allows them to infer the possible context and case law, which was not available for the ATS system.
Nevertheless, since \texttt{LMP} is subject to the legal counsellor's interpretation, this situation is not problematic as it reflects the complexity of the task.

Since we have multiple annotators, we present in \autoref{tab:annotators_res} the inter-agreement statistic of our annotators\footnote{Computed using the Prodigy inter-annotator Agreement Python package toolkit \cite{prodigy_montani_honnibal}.}, namely the percent agreement, the Krippendorff’s alpha coefficient (KAC) \cite{hayes2007answering} and the accuracy score to measure inter-annotator agreement.
We can see that annotators have a high agreement over the agreement score and accuracy for the \texttt{simplicity level} and \texttt{characterization}. 
However, the KAC of the \texttt{simplicity level} and \texttt{LMP} shows a weak agreement. 
This is because annotators D and E regularly disagree with the other three annotators. 

\begin{table}
    \centering
     \resizebox{0.49\textwidth}{!}{%
    \begin{tabular}{lccc}
    \toprule
     & \begin{tabular}[c]{@{}c@{}}Agreement (\%) ($\uparrow$)\end{tabular}& \begin{tabular}[c]{@{}c@{}}Krippendorff's $\alpha$ ($\uparrow$)\end{tabular} & \begin{tabular}[c]{@{}c@{}}Accuracy (\%) ($\uparrow$)\end{tabular}\\\midrule
    \texttt{Simplicity Level} & 57.17                              & 0.18                                     &48.11                             \\
    \texttt{Characterization} & 60.24                              & 0.55                                     & 58.05                             \\
    \texttt{Legal Meaning Preservation}  & 25.96                              & 0.10                                     & 18.48  \\
    \midrule
    Average & 47.74& 0.28 & 42.84\\
    \bottomrule                          
    \end{tabular}
    }
    \caption{Annotators inter-agreements metrics per annotation task and average. $\uparrow$ means higher is better.}
    \label{tab:annotators_res}
\end{table}

\paragraph{Final Annotation} To select the final annotation, we use a majority vote for \texttt{simplicity level} and \texttt{characterization}, and in case of ties we randomly select between the equal options. For the \texttt{LMP}, we compute an average score.

\subsection{Corpora Analysis}
\autoref{tab:datasets_statistics} presents some key statistics of FrJUDGE and the other French simplification and legal corpora introduced in \autoref{subsec:frdatasets}, where the lexical richness corresponds to the ratio of a sentence number of unique words over the overall vocabulary cardinality without removing the stop words or normalizing them \citep{van2007comparing}. 
We excluded Alector since that dataset is \href{https://alectorsite.wordpress.com/corpus/}{not available for download}.
For all corpora, we have used the latest official version on the \href{https://huggingface.co/datasets}{HuggingFace Datasets Hub}. 
All statistics were computed using SpaCy \cite{Honnibal_spaCy_Industrial-strength_Natural_2020} and exclude new lines (\verb|\n|), whitespaces and punctuations. 
We can see in \autoref{tab:datasets_statistics} that our FrJUDGE datasets' statistics, namely lexical richness and length size, are quite similar to those of the compare corpora.

\begin{table}
    \centering
    \footnotesize
    \resizebox{0.49\textwidth}{!}{%
    \begin{tabular}{@{}l|cc|cccc}
        \toprule
        & \multicolumn{2}{c|}{FrJUDGE} & \multicolumn{2}{c}{CLEAR} & \multicolumn{2}{c}{WikiLarge FR}\\
         & Complex & Simple & Complex & Simple & Complex & Simple\\\midrule
        $\#$ of sentences & 297 & 297 & 4,596 & 4,596 & 297,753 & 297,753\\
        % Vocab size & 1,776 & 1,629 & 11,008 & 10,498 & 175,582 & 147,266\\
        Lexical richness & 1.2 & 1.1 & 2.4 & 2.28 & 0.59  & 0.5\\
        % Avg $\#$ of tokens & 63.71 & 42.18 & 22.95 & 22.79 & 24.65 & 17.90 \\
        % Avg $\#$ of sent & 3.38 & 2.96 & 1.17 & 1.17 & 1.69 & 1.53 \\
        Avg sent len & 18.83 & 14.24 & 19.65 & 19.50 & 14.57 & 11.73 \\
        % Avg FKGL ($\downarrow$) & 48.89 & 68.59 & 57.65 & 58.70 & 67.06 & 73.64\\ 
    \bottomrule
    \end{tabular}%
    }
    \caption{Aggregate statistics of FrJUDGE, and the French simplification introduced in \autoref{subsec:frdatasets}.}
    \label{tab:datasets_statistics}
\end{table}

\section{\texttt{JUDGEBERT}}
\label{sec:judgebert}
We propose \texttt{JUDGEBERT}, the first supervised automatic metric for \texttt{LMP} that correlates with human judgment and passes the two sanity checks.
\texttt{JUDGEBERT} is built upon the \href{https://huggingface.co/almanach/camembertv2-base}{CamemBERT-baseV2} model \citep{antoun2024camembert20smarterfrench},
but uses a regression head instead of a classification one and feeds sentences pair into the network by concatenating them with a \verb|[SEP]| token.
CamemBERTV2-base is the smallest CamemBERTV2 model, built on the RoBERTa architecture with 112 million parameters.
It comprises 12 768-transformer layers and attention heads. \texttt{JUDGEBERT} is trained by fine-tuning the pretrained model for at most 100 epochs with an initial learning rate of $5e-5$, a patience of 5 epochs, a batch size of 16, and a linear learning rate decay as suggested by \citet{mosbach2020stability} using a 10-fold approach with a different random seed ($[42, \cdots, 51]$) to split the dataset in a 60-10-30~\% train-validation-test split and initialize the new regression attention head weights. 
% Depending on the dataset used, training takes 30 to 60 minutes using an RTX ADA 6000 GPU with 49 GO.

\section{Experimental Setup}
\label{sec:exp}
In this section, we discuss our experimental setup.
First, we discuss the automatic metrics we studied and the two sanity checks used in our experiments and finish with the training details of \texttt{JUDGEBERT}.

\subsection{Selected Metrics}
\label{subsec:metrics}
Since no automatic metric exists for preserving meaning in French or legal meaning, our experimentation builds upon previous studies on meaning preservation. Specifically, we rely on the findings of \citet{beauchemin2023meaningbert}, which suggest that Transformer-based metrics correlate better with human judgments.
Thus, for our experiments, we limited ourselves to Transformer-based metrics. We selected the following ones:

\begin{itemize}[leftmargin=*, noitemsep, topsep=0ex]
    \item \textbf{BERTScore} \citep{zhang2019bertscore} uses BERT monolingual English contextual word embeddings and computes the cosine similarity between the tokens of two sentences. It can compute precision, recall, and F1 Score over two sentences. We selected only the F1 Score since \citet{beauchemin2023meaningbert} have shown that BERTScore precision and recall scores are relatively similar for meaning preservation, and our initial experiments have shown similar results.
    \item \textbf{Sentence Transformer} \citep{reimers2019sentencebert} uses a siamese network to compare two-sentence embeddings using the cosine similarity. We selected the best monolingual English pretrained (\texttt{sentence-t5-xxl}) (SBERT) and multilingual (\texttt{distiluse-base-multilingual-cased-v1}) (SBERT-Multi) model from the \href{https://sbert.net/index.html}{official Python library}.
    \item \textbf{Coverage} was introduced by \citet{laban2021summary} to assess the meaning preservation between two texts.
    It uses a cloze test \cite{taylor1953cloze} to assess whether a monolingual English LM can fill the masked source document using a summary generated from it.
    \item \textbf{QuestEval} \citep{rebuffel2021data} is a metric designed to evaluate ATS output quality using synthetic questions and a monolingual English QnA model to respond to the generated questions using the simplification.
    The intuition is that if a simplified text conveys the same information as the source, a QnA model should be able to respond appropriately to a set of questions based on the source text.
    \item \textbf{LENS} \citep{maddela2022lens} is a trained metric for ATS quality assessment built upon monolingual English BERT. 
    \item \textbf{MeaningBERT} \citep{beauchemin2023meaningbert} is a trained metric built upon a monolingual English BERT-like model for meaning preservation between two sentences, but it does not focus on legal meaning.
\end{itemize}

\subsubsection{Nomalization}
We normalize the outputs of the different systems by decimal scaling, so those whose outputs are in $[0, 100]$ or $[0, 1]$ all line up in a $[0, 10]$ range.

\subsubsection{Semantic Capabilities}
It is essential to note that only one metric utilizes multilingual embeddings, while all the others rely on monolingual English embeddings.
Thus, none are specialized in French, while ours leverage French-specialized embeddings.
Nonetheless, our objective is to study relevant metrics and evaluate if they correlate well with human judgment, regardless of their semantic initial capabilities.

\subsection{Sanity Checks}
\label{subsec:illogical}
As per \citet{beauchemin2023meaningbert}, we also conduct two automated sanity checks as an alternative evaluation of the metrics. The checks evaluate \texttt{LMP} between identical and unrelated sentence pairs.
In these checks, the legal meaning preservation is a non-subjective measure that does not require human annotation for its assessment.
They are trivial and minimal thresholds that a good automatic \texttt{LMP} metric should be able to achieve. 
For our experiments, we compute the ratio of identical sentence pairs that score equal or greater to 99\%, and the ratio of unrelated sentence pairs that score equal or below 1\%. 
We allow a 1\% margin in each case to account for computer floating-point inaccuracy.
To generate unrelated sentences, the authors of \citet{beauchemin2023meaningbert} used GPT-2 to generate a random sentence and pair it with an unannotated sentence taken from the ASSET \cite{alva2020asset} corpus.
This approach works well for sentences that use common language. 
However, our legal corpus uses less common vocabulary than standard NLP corpus. Pairing sentences from legal documents with unrelated randomly generated sentences could make the sanity checks too trivial.
Instead, we sampled a sentence from the Québec Automobile Insurance Act \citep{loiassauto} and matched it with a sentence taken from the Québec Road Safety Code \citep{loicoderoute} that reached a maximum ROUGE-$[1, 2, \text{L}]$ and BLEU score of 0.25 and 25, respectively.
\autoref{tab:unrelated} illustrates an example of two matched sentences to illustrate how the two sentences use similar lexical vocabulary yet are unrelated.

\begin{table*}[ht!]
    \centering
    \footnotesize
    \begin{tabular}{p{\textwidth}}
        \toprule
        The persons referred to in sections 97, 99 and 100 must, at the request of a peace officer, surrender their permit for examination. \\
        \midrule
        The Minister of Revenue may, without the consent of the person concerned, communicate to the Company any information necessary for the administration of the International Registration System.\\ 
        \bottomrule
    \end{tabular}
    \caption{Example of two matched unrelated sentences (translated) randomly sampled from two legal sources. The pair reach at most a ROUGE-[1, 2, L] and BLEU scores of 0.25 and 25, respectively.}
    \label{tab:unrelated}
\end{table*}

\subsection{Training and Evaluation Datasets}
Since \texttt{JUDGEBERT} is a trainable metric, we specify the datasets used to benchmark all metrics.

\subsubsection{\texttt{JUDGEBERT} Training Datasets}
To train \texttt{JUDGEBERT}, we use FrJUDGE legal meaning human annotations, and the complex and LLM-generated simplification sentences to form a triple.
During training, we use two datasets: one using FrJUDGE 297 sentence triplets and a second that uses 594 sanity-check data augmented (DA) sentence triplets along with the FrJUDGE corpus, for a total of 891 sentence triplets.
We will refer to them as \texttt{JUDGEBERT} and \texttt{JUDGEBERT-DA}, respectively\footnote{All corpora use human-annotated sentence triplets; the data augmented corpus only adds a new sentence to the corpus.}.
We hypothesize that our data augmentation approach will improve \texttt{JUDGEBERT}'s performance on our two sanity checks, thus creating a more logical response by the metric for such cases.

\subsubsection{Evaluation Datasets}
We evaluate all selected metrics and \texttt{JUDGEBERT} on the same FrJUDGE test split during the test phase, either using the split with or without data augmentation (DA).

\subsubsection{Evaluation Metrics}
To investigate how well metrics correspond with human judgments of \texttt{LMP}, we evaluate them as machine learning models.
We use the Pearson correlation \citep{zar2005spearman} and RMSE \citep{james2013introduction} between each metric's scores and human judgment.
%\footnote{To mitigate confusion between automatic metrics \guillemet{score} studied in this work (e.g. BERTScore) and automatic metrics \guillemet{score} used to evaluate these metrics (e.g. R$^2$), we will refer to the former as \guillemet{ratings} and the latter as \guillemet{scores}.}

\subsubsection{Sanity Checks Hold-out Datasets}
To benchmark all metrics and \texttt{JUDGEBERT} on our sanity checks, we use a hold-out dataset composed of unseen sentences taken from the unused sentences in our two legal-related corpora to create an unrelated match and generate 297 related and unrelated sentences as a hold-out evaluation corpus.

\section{Metrics Ratings Analysis}
\label{sec:metrics_analysis}
In this section, we analyze the selected metrics and \texttt{JUDGEBERT} for their ability to evaluate \texttt{LMP}. 
\autoref{tab:resultats} presents the evaluation results of all metrics and our two sanity checks. For \texttt{JUDGEBERT}, we display the average score and one standard deviation. \textbf{Bolded} values are the best results per column. We also display in \autoref{ann:loss} the training and evaluation loss for \texttt{JUDGEBERT} and discuss overfitting risk.

\begin{table}[ht!]
    \centering
    \resizebox{0.5\textwidth}{!}{%
    \begin{tabular}{llcccc}
        \toprule
        DA & Metric & Pearson ($\uparrow$) & RMSE ($\downarrow$) & \% $>$ 99\% ($\uparrow$) & \% $>$ 1\% ($\uparrow$)\\
        \midrule
         & BERTScore  & 0.46 & 3.61 & \textbf{100.00} & 0.00\\
         & Coverage  & 0.19 & 2.82  & 0.00 & 0.00 \\
         & LENS  & 0.38 & 2.57  & 0.00 & 0.67\\
         False & MeaningBERT  & 0.17 & 3.51  & \textbf{100.00} & 0.67 \\
         & QuestEval  & -0.05 & 2.99  & 0.00 & 0.00\\
         & SBERT  & 0.13 & 3.25  & \textbf{100.00} & 0.00 \\
         & SBERT-Multi & 0.06 & 3.35 & 0.00 & 0.00 \\
         & \texttt{JUDGEBERT} & 0.74 $\pm$ 0.02 & 1.72 $\pm$ 0.10  & 0.00 & 0.00\\
        \midrule
         & BERTScore  & 0.94 & 5.09  & \textbf{100.00} & 0.00\\
         & Coverage  & 0.90 & 2.20  & 0.00 & 0.00\\
         & LENS  & 0.56 & 3.87  & 0.00 & 0.67\\
         True & MeaningBERT  & 0.81 & 3.98  & \textbf{100.00} & 0.67 \\
         & QuestEval  & 0.68 & 3.82  & 0.00 & 0.00\\
         & SBERT  & 0.92 & 2.84  & \textbf{100.00} & 0.00\\
         & SBERT-Multi & 0.90 & 2.39 & \textbf{100.00} & 0.00 \\
         & \texttt{JUDGEBERT-DA} & \textbf{0.97 $\pm$ 0.00} & \textbf{1.01 $\pm$ 0.07}  & \textbf{100.00} & \textbf{100.00}\\
    \bottomrule
    \end{tabular}
    }
    \caption{Results of the selected metrics and \texttt{JUDGEBERT} trained with or without data augmentation (DA). We also present one standard deviation for trained models. \textbf{Bolded} values are the best results overall. $\uparrow$ means higher is better, while $\downarrow$ mean otherwise.}
    \label{tab:resultats}
\end{table}

\subsection{Metrics Ratings and Human Judgments}
First, we can see in \autoref{tab:resultats} that Pearson correlation scores vary greatly between metrics, with an average correlation between $[-0.05, 0.74]$ and $[0.56, 0.97]$, with and without DA, respectively.
This shows that not all metrics are suitable for our task.
Indeed, we can see that most selected metrics have a low to moderate degree of correlation with human judgment, with BERTScore reaching the second-highest score. 
We can also see that all metrics achieve a higher correlation with human judgment when DA is introduced, meaning they can, to a certain degree, be compliant with our two sanity checks.
Furthermore, \texttt{JUDGEBERT} achieves the highest correlation with human judgment, with a near-perfect correlation when trained with DA.

On the other hand, we can see that all selected metrics achieve poor performance on RMSE, higher than \texttt{JUDGEBERT} with and without DA. Since our labels are on a 10-point Likert scale, the RMSE corresponds to the number of levels of difference between the model's output and human judgement. Our results thus demonstrate that the selected metrics are, on average, very different from human judgments.
Furthermore, since we want to assess \texttt{LMP}, the impact of a \guillemet{close enough} score differs depending on whether the score is higher or lower than the human evaluation.
Indeed, in practice, a system that undershoots human judgment is simply strict in the simplifications it accepts, but one that overshoots human judgment is unacceptably permissive of bad simplifications.
Thus, we present in \autoref{tab:analyse_1} the percentage of predictions with a higher output than the human judgment on the corpus without DA. For all metrics, except our \texttt{JUDGBERT} models, the output score is regularly higher than human judgments. It shows that other metrics are inadequate for \texttt{LMP}.

\begin{table}[ht!]
    \centering
    \footnotesize
    \begin{tabular}{lc}
        \toprule
        Metric & \% > labels ($\downarrow$)\\
        \midrule
         BERTScore  & 82.22 \\
         Coverage  & 27.78 \\
         LENS  & 30.00 \\
         MeaningBERT  & 82.22 \\
         QuestEval  & 27.78 \\
         SBERT  & 76.67 \\
         SBERT-Multi & 77.78\\
         \texttt{JUDGEBERT} & \textbf{0.00 $\pm$ 0.00}\\
         \texttt{JUDGEBERT-DA} & \textbf{0.00 $\pm$ 0.00}\\
    \bottomrule
    \end{tabular}
    \caption{Percentage of predictions with a higher rating than the human judgments of the selected metrics and \texttt{JUDGEBERT} on the test set without DA. \textbf{Bolded} values are the best results. $\uparrow$ means higher is better.}
    \label{tab:analyse_1}
\end{table}

\subsection{Metrics Sanity Checks}
We can see in \autoref{tab:resultats} that only three metrics always return the expected value of 100\% (e.g., 99\% to account for rounding error) when comparing two identical sentences: BERTScore, SBERT, SBERT-Multi and MeaningBERT. 
These results are expected for all metrics, as BERTScore employs an algorithm that returns a perfect score when the two texts are identical. MeaningBERT was trained to do so, and SBERT and SBERT-Multi both use cosine similarity between embeddings to compute the similarity. 
Thus, two similar sentences will return the same vectors.

On the other hand, none of the metrics achieve a perfect performance on the second check. This poor performance is similar to the results observed by 
\citet{beauchemin2023meaningbert}. 
These authors hypothesize that BERT-like metrics that use contextualized embeddings can hallucinate connections and common meaning between the two sentence vectors even when none exist, thus returning a non-zero rating.
This is likely our case since we use unrelated sentences with a similar legal lexicon but from two different sources.
It shows that without proper legal knowledge, unrelated sentences can seem similar. 
This is a significant limitation of existing metrics in our case: 
since we evaluate \texttt{LMP}, generating a score different from zero for two completely unrelated sentences significantly reduces a metric's credibility for a legal counsellor.

Finally, we can see that with DA, \texttt{JUDGEBERT-DA} can pass both sanity checks. It shows that an LM cannot capture the coherent logic embedded in our sanity checks without being given proper examples. 

\section{Conclusion and Future Work}
\label{sec:conclusion}
This paper proposes a new metric to assess legal meaning preservation between two legal sentences, specifically in the context of text simplification. However, our metric could also be used for other tasks. 
We also proposed FrJUDGE, a new legal meaning judgment dataset consisting of 297 human-annotated sentences taken from French insurance legal documents.
To demonstrate its quality and versatility, we compared our work against a set of Transformer-based metrics in the literature applied to FrJUDGE. Further, we applied two automatic sanity checks to evaluate meaning preservation between identical and unrelated sentences. 
In future work, we aim to study how \texttt{JUDGEBERT} generalizes to other languages and tasks. 
We also aim to increase FrJUDGE's size by including other insurance products, such as group insurance.
Finally, we also want to expand FrJUDGE's size by including pieces of text that are not jurisdiction-specific, such as French versions of arbitration and mediation clauses, which are subject to international conventions and not region-specific.

\section*{Limitations}
All the sentences included in FrJUDGE have been extracted from para-governmental official sources. Therefore, they are guaranteed to be meaningful, making FrJUDGE a challenging dataset.
However, text instances are relatively short and are analyzed by legal experts outside their context; thus, this differs from how contracts are typically analyzed (i.e. as a whole), and the application of a contract depends to a large extent on the facts \cite{cardon2020french}.
Such an approach, which contextualizes the overall document for text simplification, is more coherent with the recent work of \citet{agrawal2024text}.
However, doing such an evaluation would be more costly and complex to orchestrate. Nevertheless, such approaches have been conducted with corpus such as CUAD \cite{hendrycks2021cuad}. To generate such a complex dataset, the cost of CUAD is estimated to be in the millions of dollars, whereas our annotation budget was USD 5,000.

\texttt{JUDGEBERT} has been trained on a relatively small dataset (i.e. FrJUDGE) for such a large model, and it has only seen a subset of all types of legal documents (namely insurance text).
Moreover, our trained models were not tested with an out-of-domain (OOD) split to assert any overfitting risk. Thus, JUDGEBERT may have overfitted our training splits.
However, we hope that the NLP community's interest in this work will lead to the development of robust metrics to assess the legal aspect of deep learning models.

As shown in \autoref{subsec:annotation_res}, assessing the \texttt{legal meaning precision} of text is complex and is subject to interpretation. Interpreting whether or not a reformulation of a text conveys the same legal meaning will always be an approximation, and the only real complete test would be to discuss it in tribunals. However, such assessments are nearly impossible on a large scale. Thus, we argue that our approach can give insightful information to any legal practitioner on the overall \texttt{legal meaning precision} of a legal TS rather than a complete juridical analysis. However, our approach should not be considered legal advice, and \texttt{JUDGEBERT} should not be considered a comprehensive legal expert. 

Finally, the metrics we selected in our study (\autoref{subsec:metrics}) are mostly English-based approaches, yet we applied them to French text, which may give our French-based approach a potentially unfair advantage. Thus, our study cannot conclude whether some of these metrics are irrelevant to the preservation of legal meaning in English. 

\section*{Ethical Considerations}
FrJUDGE may serve as training data for French legal classifiers \cite{batra-etal-2021-building}, as an expert source for text to structure legal expert systems \cite{sileno2023text}, or for training specialized LLM in French \cite{douka2021juribert, 10.1145/3462757.3466147}, which may benefit the quality of generated texts in the legal field \cite{tan2023chatgpt, kapoor2024promises}. 
Our corpus can be used to enhance online legal resources, providing laypeople with access to juridical services \cite{hagan2023good, kapoor2024promises}.
We acknowledge that such text generation progress could lead to the misuse of LLMs for malicious purposes, such as legal disinformation or harmful text generation \cite{weidinger2021ethical, bender2021dangers, hagan2023good, kapoor2024promises}. 
However, our corpus can also be used for training adversarial defence systems against such misuses and to train artificial text detection models, \cite{lewis-white-2023-mitigating, kumar-etal-2023-mitigating}.

\texttt{JUDGEBERT} may serve as a metric for evaluating LLMs in the legal and insurance domains. 
Legal documents are more challenging to read than typical documents; simplifying these documents can prove to be costly, so assessing the quality of legal documents is also costly \cite{hendrycks2021cuad}.
We acknowledge that using trained metrics could lead to misuse and blind faith in users who trust such metrics.
Nevertheless, our metric can be further improved to increase laypersons' access to proper legal expertise.

\section*{Acknowledgements}
This research was made possible thanks to the support of a Canadian insurance company, NSERC research grant RDCPJ 537198-18 and FRQNT doctoral research grant. We thank the reviewers for their comments regarding our work as well as our colleagues from the \textit{Groupe de recherche en apprentissage automatique de l'Université Laval} who peer-reviewed our manuscript. 
Lastly, we wish to thank the annotators who evaluated the generations.

\bibliography{anthology,custom}
\bibliographystyle{acl_natbib}

\appendix

\section{Automatic Text Simplification Prompt and Generation Parameters}
\label{ann:prompt}
The \autoref{fig:promptFR} presents the French prompt used for generating the TS, along with the automatic translation in English using DeepL machine translator\footnote{\href{https://www.deepl.com/translator}{https://www.deepl.com/translator}} (\autoref{fig:promptEN} for the non-French reader); \verb|{input}| is the placeholder for the complex sentence.
It is based on \citet{kew2023bless} basic zero-shot prompt, but we made the following two modifications:

\begin{enumerate}[leftmargin=*, noitemsep]
    \item Manual translation from English to French;
    \item Add specification in uppercase to respond in French since uppercase capitalization has increased importance to an instruction \cite{ozdemir2023quick, tornberg2024best, hu2024improving}.
\end{enumerate}

\begin{figure}[ht!]
        \begin{subfigure}[b]{\linewidth}
        \footnotesize
        \centering
            \begin{tikzpicture}[scale=1, every node/.style={transform shape}]
            
            % Blue rectangle for "Prompt"
            \node[rectangle, rounded corners, draw=tgb1, fill=tgb1!80, text width=0.9\linewidth, align=left, inner sep=1.2ex] (prompt) {Réécris la phrase complexe à l'aide d'une ou plusieurs phrases simples. Conserve le même sens, mais simplifie-le. RÉPONDS EN FRANÇAIS!};
            
            % Yellow rectangle for "Input"
            \node[rectangle, rounded corners, draw=tgb4, fill=tgb4, below=0.1cm of prompt, text width=0.9\linewidth, align=left, inner sep=1.2ex] (input) {Complexe: \{input\}.};
            \end{tikzpicture}
            \caption{Basic zero-shot prompt adapted from \citet{kew2023bless} followed by the input sentence to be simplified.}
            \label{fig:promptFR}
        \end{subfigure}
        
        \begin{subfigure}[b]{\linewidth}
        \centering
            \begin{tikzpicture}[scale=1, every node/.style={transform shape}]
    
            % Blue rectangle for "Prompt"
            \node[rectangle, rounded corners, draw=tgb1, fill=tgb1!80, text width=0.9\linewidth, align=left, inner sep=1.2ex] (prompt) {Rewrite the complex sentence with simple sentence(s). Keep the meaning the same, but make it simpler. ANSWER IN FRENCH!};
            
            % Yellow rectangle for "Input"
            \node[rectangle, rounded corners, draw=tgb4, fill=tgb4, below=0.1cm of prompt, text width=0.9\linewidth, align=left, inner sep=1.2ex] (input) {Complex: \{input\}.};
            \end{tikzpicture}
            \caption{Translation of the prompt presented in \autoref{fig:promptFR}.}
            \label{fig:promptEN}
        \end{subfigure}
    \caption{Prompts used for LLM text simplification. 
    \hlc{tgb1!80}{Bleu} boxes contain the task instructions. \hlc{tgb4}{Yellow} boxes contain the prefix for the model to continue.}
    \label{fig:prompts}
\end{figure}

\autoref{tab:parameters} present the OpenAI generation parameters used for generating the simplification. We have used the same parameters as per \citet{kew2023bless}. The cost of generating all 312 simplifications was less than 5 USD.

\begin{table}[ht!]
    \centering
    \begin{tabular}{lc}
    \toprule
    Parameter         & Value                  \\\midrule
    Model name        & gpt-4-turbo-2024-04-09 \\
    Max new tokens    & 100                    \\
    Temperature       & 1.0                    \\
    Top K             & 0.9                    \\
    Frequency penalty & 0.0                    \\
    Presence penalty  & 0.0                   \\\bottomrule
    \end{tabular}
    \caption{OpenAI generation parameters used for generating the simplification}
    \label{tab:parameters}
\end{table}

\section{Generation Examples}
\label{ann:generation}
The \autoref{fig:generations} presents examples of original text in French (\hlc{tgb2}{cyan}) along with the simplification (\hlc{tgb5}{pink}) made from the GPT4 model using the zero-shot simplification prompt and generation parameters as presented in \autoref{ann:prompt}, and their respective automatic translation in English using DeepL machine translator (\hlc{tgb6!80}{purple}, \hlc{brown}{brown} respectively).

\begin{figure}[ht!]
    \begin{subfigure}[b]{\linewidth}
        \footnotesize
        \centering
        \begin{tikzpicture}[scale=1, every node/.style={transform shape}]
        
        % orange rectangle for "original"
        \node[rectangle, rounded corners, draw=tgb2, fill=tgb2, text width=0.9\linewidth, align=left, inner sep=1.2ex] (prompt) {Franchise, Il s’agit d’un montant restant à votre charge en cas de sinistre. Ce montant est stipulé aux Conditions particulières.};
        
        % green rectangle for "simplification"
        \node[rectangle, rounded corners, draw=tgb5, fill=tgb5, below=0.1cm of prompt, text width=0.9\linewidth, align=left, inner sep=1.2ex] (input) {Franchise est le montant que vous payez en cas de sinistre. Ce montant est indiqué dans les Conditions particulières.};

        \node[rectangle, rounded corners, draw=tgb6, fill=tgb6!80, below=0.5cm of input, text width=0.9\linewidth, align=left, inner sep=1.2ex] (original) {Deductible: This is the amount you must pay in the event of a claim. This amount is stipulated in the Declarations.};

        \node[rectangle, rounded corners, draw=brown, fill=brown, below=0.1cm of original, text width=0.9\linewidth, align=left, inner sep=1.2ex] (input) {Deductible is the amount you pay in the event of a claim. This amount is indicated in the Declarations.};
        \end{tikzpicture}
        \caption{Example of a generation based on the deductible definition from the Quebec automotive insurance form.}
        \label{fig:generationsa}
    \end{subfigure}

    \begin{subfigure}[b]{\linewidth}
        \centering
        \begin{tikzpicture}[scale=1, every node/.style={transform shape}]
        
        % orange rectangle for "original"
        \node[rectangle, rounded corners, draw=tgb2, fill=tgb2, text width=0.9\linewidth, align=left, inner sep=1.2ex] (prompt) {Aux conditions ci-après, l’alinéa de l’exclusion no 15 est supprimé en ce qui concerne la garantie A de la première partie - Assurance de vos biens, de votre contrat.};
        
        % green rectangle for "simplification"
        \node[rectangle, rounded corners, draw=tgb5, fill=tgb5, below=0.1cm of prompt, text width=0.9\linewidth, align=left, inner sep=1.2ex] (input) {L'alinea de l'exclusion no 15 est supprimé de la garantie A de la première partie - Assurance de vos biens, de votre contrat, sous certaines conditions.};

        \node[rectangle, rounded corners, draw=tgb6, fill=tgb6!80, below=0.5cm of input, text width=0.9\linewidth, align=left, inner sep=1.2ex] (original) {Under the following conditions, the paragraph of exclusion no. 15 is deleted with respect to Coverage A of Part I - Property Insurance of your contract.};

        \node[rectangle, rounded corners, draw=brown, fill=brown, below=0.1cm of original, text width=0.9\linewidth, align=left, inner sep=1.2ex] (input) {Under certain conditions, the paragraph of exclusion no. 15 is deleted from coverage A of the first part of your contract - Insurance of your property.};
        \end{tikzpicture}
        \caption{Example of a generation based on the deductible definition from the Quebec principal residency insurance form.}
        \label{fig:generationsb}
    \end{subfigure}
    \caption{Examples of a generation using GPT4. 
    \hlc{tgb2!80}{Cyan} boxes contain the original French text, and \hlc{tgb6!80}{purple} boxes contain the automatic translation of the original text in English. \hlc{tgb5}{Pink} boxes contain the simplification generation in French, and \hlc{brown}{brown} boxes contain the automatic translation of the simplified generation in English.}
    \label{fig:generations}
\end{figure}

\section{Annotation Interface}
\label{ann:clf_annotation}

\begin{figure*}
    \centering
    \includegraphics[width=1\textwidth]{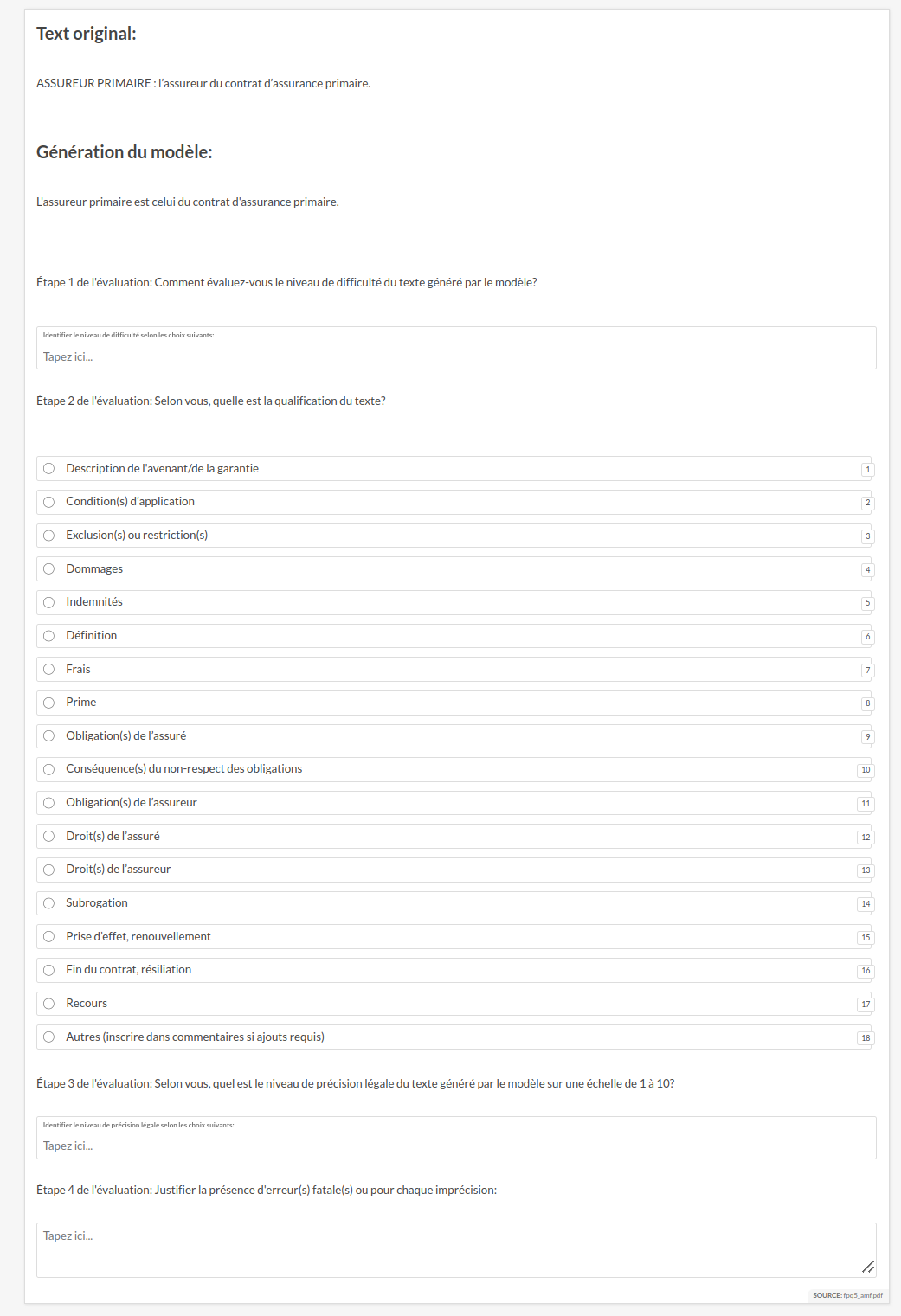}
    \caption{The Prodigy annotation interface (in French) used by the annotators to evaluate the instance generated by an ATS system.}
    \label{fig:clf_annotation}
\end{figure*}

The \autoref{fig:clf_annotation} presents the evaluation interface used by our annotators (in French). It is a custom adaptation of the Prodigy annotation tool \cite{prodigy_montani_honnibal}. 

\section{Characterization Class}
\label{ann:charac}
In this section, we detail the characterization class used by our annotator. For each, we present the characterization in French, an automatic English translation, and a brief description in English. All description were taken from \citet{assurance_book}. 

% https://docs.google.com/document/d/1CMjx5uYUxmnCLNqrtAKBHwuPpPA0N8Fc/edit
% https://docs.google.com/document/u/2/d/1RpH7VLpvqGEp2-D_qNIkNzvWdyzOtI25GDPlRaGlc5k/edit?tab=t.0
\begin{enumerate}[leftmargin=*, noitemsep]
    \item \textbf{Description of endorsement (\textit{Description de l’avenant})}: These are appendices that modify the basic insurance contract, such as the \guillemet{replacement cost} coverage endorsement. The text of the endorsement takes precedence over the general text of the insurance policy.
    \item \textbf{Conditions of application (\textit{Conditions d’applications})}: Refers to the general conditions of application of either an insurance contract or endorsements. For example, \guillemet{subject to risk acceptance}.
    \item \textbf{Exclusions or restrictions (\textit{Exclusions ou restrictions})}: Refers to the general exclusions or restrictions that can apply to the insurance contract or the endorsements. For example, \guillemet{exclusions of replacement value} or \guillemet{exclusions of nuclear damage}.
    \item \textbf{Damage (value of, calculation of and description of) (\textit{Dommages (valeur des, calcul des et description des)})}:
    Refers to the mechanism and principles to assess the value of the damage after an incident.
    \item \textbf{Indemnities (indemnities payable, indemnity per replacement, calculation of value of, amount of insurance and indemnity process) (\textit{Indemnités (indemnités payables, indemnité par remplacement, calcul de la valeur des, montant d’assurance et processus d’indemnisation)})}: Refers to the mechanism and principles to assess the indemnities amount payable to an insuree, the principles of a replacement of the damaged property, the methodology to evaluate the value of the damage properties and indemnisation process along with the resolution in case of disagreement.
    \item \textbf{Definition (\textit{Définition})}:
    Refers to definitions of specific terms in the contract, endorsements or other legal elements. For example, \guillemet{definition of deductible}.
    \item \textbf{Expenses (reimbursement and assumption of costs) (\textit{Frais (remboursement et prise en charge des)})}:
    Refers to the principles of expense reimbursement in case of an insured incident, such as towing the insured car or expenses to minimize damage.
    \item \textbf{Premium (payment and reimbursement of) (\textit{Prime (paiement de et remboursement de)})}: Refers to premium details such as the amount, how and when to pay it, and the reimbursement terms.  
    \item \textbf{Obligations of the insured (obligation and formal commitment) (\textit{Obligations de l’assuré (obligation et engagement formel)})}:
    Refers to the insuree's obligations to be executed during the duration of the contract. For example, \guillemet{Risk aggravation declaration}.
    \item \textbf{Consequences of non-compliance (\textit{Conséquences du non-respect des obligations})}:
    Refers to the consequences of non-compliance to the insuree or insurer engagements, such as indemnity reduction or legal actions of the insurer against its insuree (e.g. false declaration).
    \item \textbf{Insurer's obligations (\textit{Obligations de l’assureur})}:
     Refers to the insurer's obligations to be executed during the contract duration. For example, \guillemet{insurer's obligation to inform and advise the insured}.
    \item \textbf{Insured's rights (including waiver of rights) (\textit{Droits de l’assuré (incluant la renonciation aux droits)})}:
    Refers to the rights of the insured regarding the insurance contract, such as the right of renewal and representation.
    \item \textbf{Insurer's rights (including waiver of rights) (\textit{Droits de l’assureur (incluant la renonciation aux droits)})}:
    Refers to the insurer's right regarding the insurance contract, such as the right to refuse coverage. 
    \item \textbf{Subrogation (and exceptions to subrogation) (\textit{Subrogation (et exceptions à la subrogation)})}:
    Refers to a specific right of the insuree and insured called subrogation right that defines the right of the insuree to transfer all its rights over an incident to the insurer. The insurer will represent the right of the insuree and protect the insuree and insurer interest.
    \item \textbf{Effective date and renewal (\textit{Prise d’effet et renouvellement})}:
    Refers to the effective date and renewal of the insurance contract.
    \item \textbf{End of contract and termination (\textit{Fin du contrat et résiliation})}:
    Refers to the effective end date and the termination conditions of the insurance contract.
    \item \textbf{Legal recourse (dispute resolution, action, representation mandate, arbitration, etc.) (\textit{Recours (règlement de différend, action, mandat de représentation, arbitrage, etc.)})}: Refers to the mechanisms for resolving legal disputes and institutions to which policyholders can refer. For example, to the regulatory body (i.e. AMF in Quebec).
    \item \textbf{Others (\textit{Autres})}: Class use when the sentence does not apply to any of the 17 previous classes. When annotators use these cases, we ask them to elaborate on why none of the earlier classes were appropriate. 
\end{enumerate}

\section{Training Loss}
\label{ann:loss}
\begin{figure*}[ht!]
    \centering
    \begin{subfigure}[b]{0.49\linewidth}
        \centering
        \includegraphics[width=\linewidth]{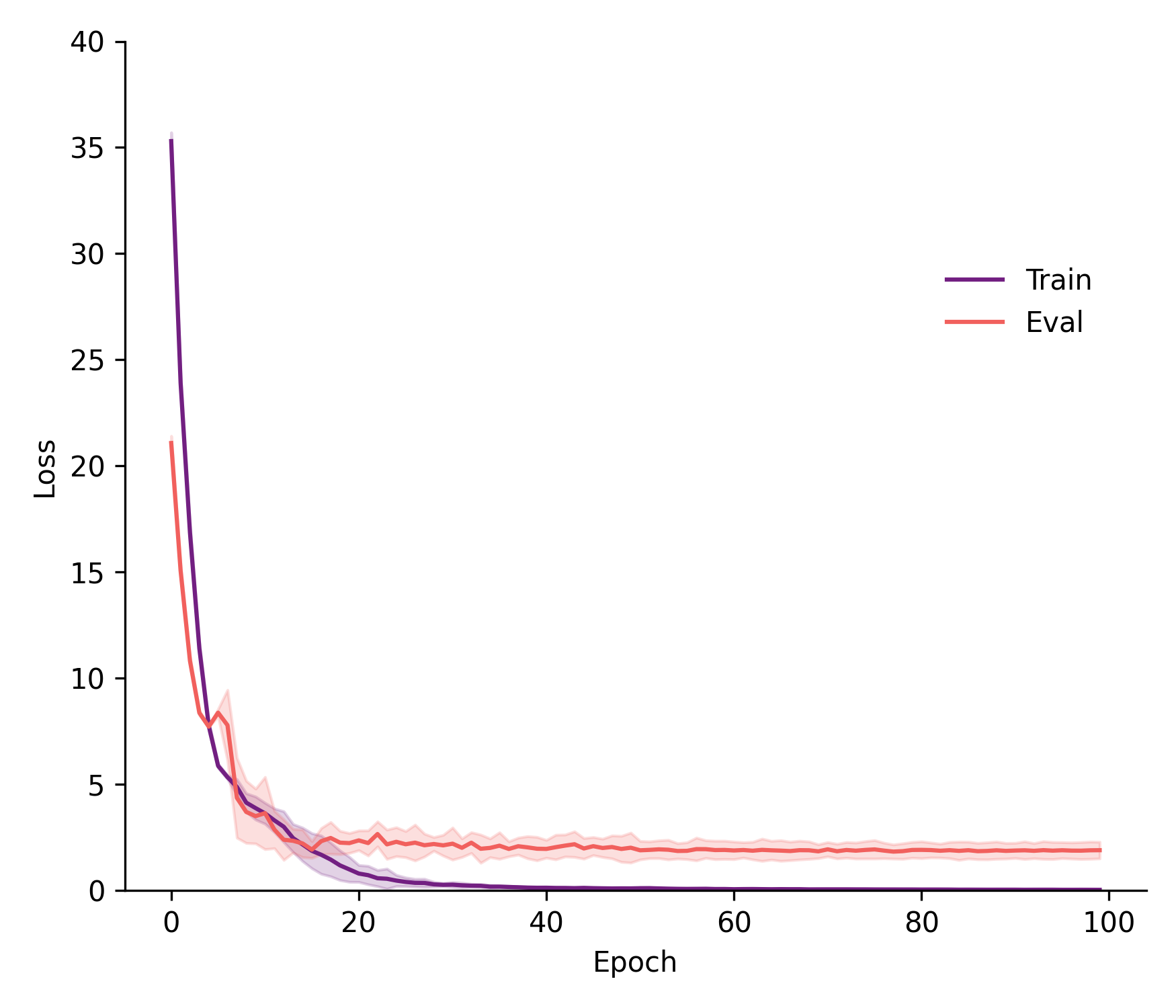}
        \caption{Training and evaluation loss for \texttt{JUDGEBERT} over the 100 epochs.}
        \label{subfig:loss-noda}
    \end{subfigure}
    \begin{subfigure}[b]{0.49\linewidth}
        \centering
        \includegraphics[width=\linewidth]{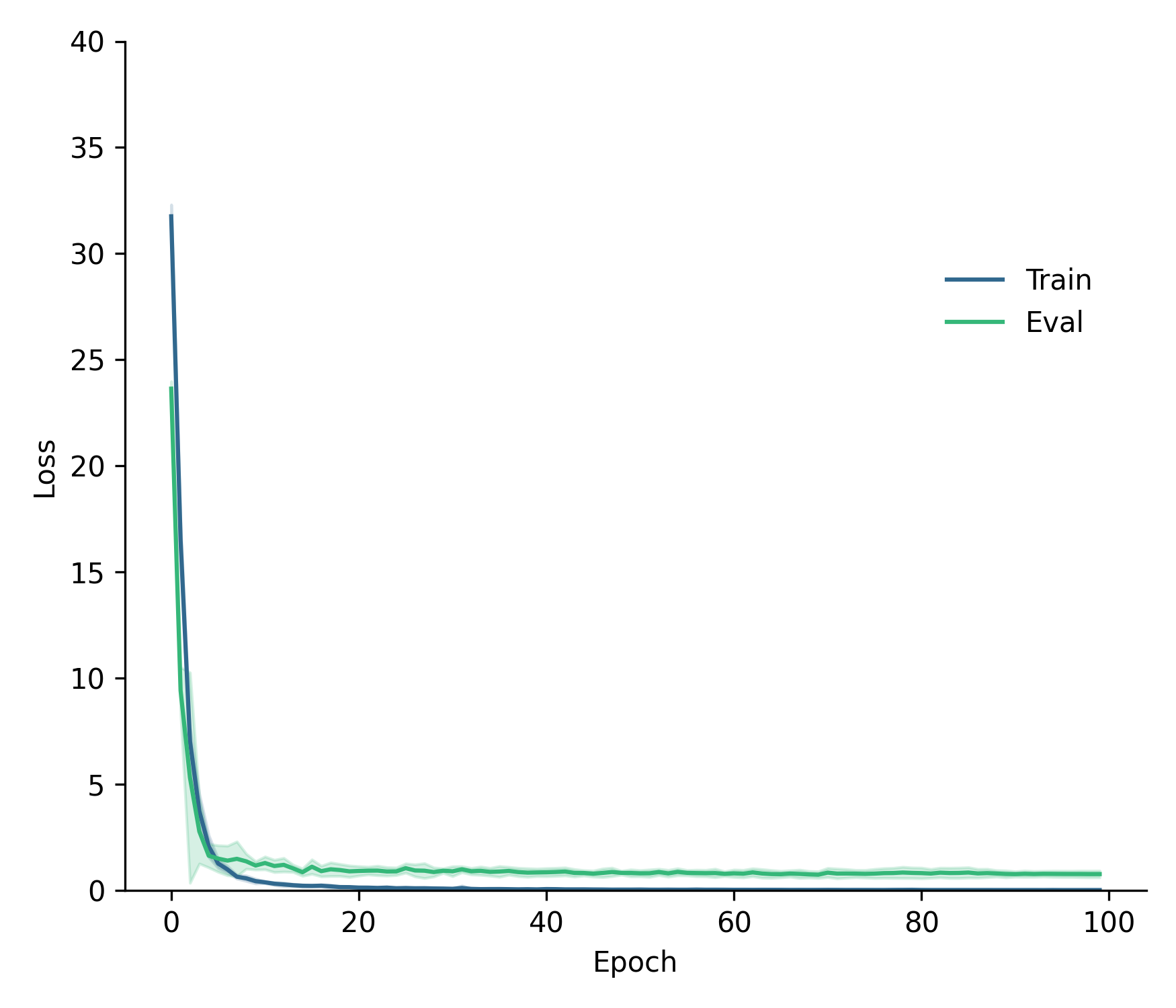}
        \caption{Training and evaluation loss for \texttt{JUDGEBERT-DA} over the 100 epochs.}
        \label{subfig:loss-da}
    \end{subfigure}
    \caption{Training and evaluation loss}
    \label{fig:loss}
\end{figure*}

In this section, we present the training and evaluation loss for trained metrics in \autoref{fig:loss}. We can see in \autoref{subfig:loss-noda} that the loss reaches a plateau after 60 epochs, resulting in a wide gap between the training and evaluation loss. It indicates potential overfitting for the \texttt{JUDGEBERT} model.
However, as shown in \autoref{subfig:loss-da}, \texttt{JUDGEBERT-DA} training and evaluation gap is smaller and tends to slowly decrease over time, meaning a lower risk of overfitting.

\input{datasheet}

\end{document}

%% file: fig/difficulty_level.pgf
%% Creator: Matplotlib, PGF backend
%%
%% To include the figure in your LaTeX document, write
%%   \input{<filename>.pgf}
%%
%% Make sure the required packages are loaded in your preamble
%%   \usepackage{pgf}
%%
%% Also ensure that all the required font packages are loaded; for instance,
%% the lmodern package is sometimes necessary when using math font.
%%   \usepackage{lmodern}
%%
%% Figures using additional raster images can only be included by \input if
%% they are in the same directory as the main LaTeX file. For loading figures
%% from other directories you can use the `import` package
%%   \usepackage{import}
%%
%% and then include the figures with
%%   \import{<path to file>}{<filename>.pgf}
%%
%% Matplotlib used the following preamble
%%   
%%   \usepackage{fontspec}
%%   \makeatletter\@ifpackageloaded{underscore}{}{\usepackage[strings]{underscore}}\makeatother
%%
\begingroup%
\makeatletter%
\begin{pgfpicture}%
\pgfpathrectangle{\pgfpointorigin}{\pgfqpoint{6.400000in}{4.800000in}}%
\pgfusepath{use as bounding box, clip}%
\begin{pgfscope}%
\pgfsetbuttcap%
\pgfsetmiterjoin%
\definecolor{currentfill}{rgb}{1.000000,1.000000,1.000000}%
\pgfsetfillcolor{currentfill}%
\pgfsetlinewidth{0.000000pt}%
\definecolor{currentstroke}{rgb}{1.000000,1.000000,1.000000}%
\pgfsetstrokecolor{currentstroke}%
\pgfsetdash{}{0pt}%
\pgfpathmoveto{\pgfqpoint{0.000000in}{0.000000in}}%
\pgfpathlineto{\pgfqpoint{6.400000in}{0.000000in}}%
\pgfpathlineto{\pgfqpoint{6.400000in}{4.800000in}}%
\pgfpathlineto{\pgfqpoint{0.000000in}{4.800000in}}%
\pgfpathlineto{\pgfqpoint{0.000000in}{0.000000in}}%
\pgfpathclose%
\pgfusepath{fill}%
\end{pgfscope}%
\begin{pgfscope}%
\pgfsetbuttcap%
\pgfsetmiterjoin%
\definecolor{currentfill}{rgb}{1.000000,1.000000,1.000000}%
\pgfsetfillcolor{currentfill}%
\pgfsetlinewidth{0.000000pt}%
\definecolor{currentstroke}{rgb}{0.000000,0.000000,0.000000}%
\pgfsetstrokecolor{currentstroke}%
\pgfsetstrokeopacity{0.000000}%
\pgfsetdash{}{0pt}%
\pgfpathmoveto{\pgfqpoint{0.634445in}{0.372083in}}%
\pgfpathlineto{\pgfqpoint{6.250000in}{0.372083in}}%
\pgfpathlineto{\pgfqpoint{6.250000in}{4.650000in}}%
\pgfpathlineto{\pgfqpoint{0.634445in}{4.650000in}}%
\pgfpathlineto{\pgfqpoint{0.634445in}{0.372083in}}%
\pgfpathclose%
\pgfusepath{fill}%
\end{pgfscope}%
\begin{pgfscope}%
\pgfpathrectangle{\pgfqpoint{0.634445in}{0.372083in}}{\pgfqpoint{5.615555in}{4.277917in}}%
\pgfusepath{clip}%
\pgfsetbuttcap%
\pgfsetmiterjoin%
\definecolor{currentfill}{rgb}{0.121569,0.466667,0.705882}%
\pgfsetfillcolor{currentfill}%
\pgfsetlinewidth{0.000000pt}%
\definecolor{currentstroke}{rgb}{0.000000,0.000000,0.000000}%
\pgfsetstrokecolor{currentstroke}%
\pgfsetstrokeopacity{0.000000}%
\pgfsetdash{}{0pt}%
\pgfpathmoveto{\pgfqpoint{0.985417in}{0.372083in}}%
\pgfpathlineto{\pgfqpoint{1.125806in}{0.372083in}}%
\pgfpathlineto{\pgfqpoint{1.125806in}{3.697010in}}%
\pgfpathlineto{\pgfqpoint{0.985417in}{3.697010in}}%
\pgfpathlineto{\pgfqpoint{0.985417in}{0.372083in}}%
\pgfpathclose%
\pgfusepath{fill}%
\end{pgfscope}%
\begin{pgfscope}%
\pgfpathrectangle{\pgfqpoint{0.634445in}{0.372083in}}{\pgfqpoint{5.615555in}{4.277917in}}%
\pgfusepath{clip}%
\pgfsetbuttcap%
\pgfsetmiterjoin%
\definecolor{currentfill}{rgb}{0.121569,0.466667,0.705882}%
\pgfsetfillcolor{currentfill}%
\pgfsetlinewidth{0.000000pt}%
\definecolor{currentstroke}{rgb}{0.000000,0.000000,0.000000}%
\pgfsetstrokecolor{currentstroke}%
\pgfsetstrokeopacity{0.000000}%
\pgfsetdash{}{0pt}%
\pgfpathmoveto{\pgfqpoint{2.389306in}{0.372083in}}%
\pgfpathlineto{\pgfqpoint{2.529695in}{0.372083in}}%
\pgfpathlineto{\pgfqpoint{2.529695in}{1.230632in}}%
\pgfpathlineto{\pgfqpoint{2.389306in}{1.230632in}}%
\pgfpathlineto{\pgfqpoint{2.389306in}{0.372083in}}%
\pgfpathclose%
\pgfusepath{fill}%
\end{pgfscope}%
\begin{pgfscope}%
\pgfpathrectangle{\pgfqpoint{0.634445in}{0.372083in}}{\pgfqpoint{5.615555in}{4.277917in}}%
\pgfusepath{clip}%
\pgfsetbuttcap%
\pgfsetmiterjoin%
\definecolor{currentfill}{rgb}{0.121569,0.466667,0.705882}%
\pgfsetfillcolor{currentfill}%
\pgfsetlinewidth{0.000000pt}%
\definecolor{currentstroke}{rgb}{0.000000,0.000000,0.000000}%
\pgfsetstrokecolor{currentstroke}%
\pgfsetstrokeopacity{0.000000}%
\pgfsetdash{}{0pt}%
\pgfpathmoveto{\pgfqpoint{3.793195in}{0.372083in}}%
\pgfpathlineto{\pgfqpoint{3.933584in}{0.372083in}}%
\pgfpathlineto{\pgfqpoint{3.933584in}{0.731113in}}%
\pgfpathlineto{\pgfqpoint{3.793195in}{0.731113in}}%
\pgfpathlineto{\pgfqpoint{3.793195in}{0.372083in}}%
\pgfpathclose%
\pgfusepath{fill}%
\end{pgfscope}%
\begin{pgfscope}%
\pgfpathrectangle{\pgfqpoint{0.634445in}{0.372083in}}{\pgfqpoint{5.615555in}{4.277917in}}%
\pgfusepath{clip}%
\pgfsetbuttcap%
\pgfsetmiterjoin%
\definecolor{currentfill}{rgb}{0.121569,0.466667,0.705882}%
\pgfsetfillcolor{currentfill}%
\pgfsetlinewidth{0.000000pt}%
\definecolor{currentstroke}{rgb}{0.000000,0.000000,0.000000}%
\pgfsetstrokecolor{currentstroke}%
\pgfsetstrokeopacity{0.000000}%
\pgfsetdash{}{0pt}%
\pgfpathmoveto{\pgfqpoint{5.197083in}{0.372083in}}%
\pgfpathlineto{\pgfqpoint{5.337472in}{0.372083in}}%
\pgfpathlineto{\pgfqpoint{5.337472in}{0.465743in}}%
\pgfpathlineto{\pgfqpoint{5.197083in}{0.465743in}}%
\pgfpathlineto{\pgfqpoint{5.197083in}{0.372083in}}%
\pgfpathclose%
\pgfusepath{fill}%
\end{pgfscope}%
\begin{pgfscope}%
\pgfpathrectangle{\pgfqpoint{0.634445in}{0.372083in}}{\pgfqpoint{5.615555in}{4.277917in}}%
\pgfusepath{clip}%
\pgfsetbuttcap%
\pgfsetmiterjoin%
\definecolor{currentfill}{rgb}{1.000000,0.498039,0.054902}%
\pgfsetfillcolor{currentfill}%
\pgfsetlinewidth{0.000000pt}%
\definecolor{currentstroke}{rgb}{0.000000,0.000000,0.000000}%
\pgfsetstrokecolor{currentstroke}%
\pgfsetstrokeopacity{0.000000}%
\pgfsetdash{}{0pt}%
\pgfpathmoveto{\pgfqpoint{1.125806in}{0.372083in}}%
\pgfpathlineto{\pgfqpoint{1.266195in}{0.372083in}}%
\pgfpathlineto{\pgfqpoint{1.266195in}{3.150661in}}%
\pgfpathlineto{\pgfqpoint{1.125806in}{3.150661in}}%
\pgfpathlineto{\pgfqpoint{1.125806in}{0.372083in}}%
\pgfpathclose%
\pgfusepath{fill}%
\end{pgfscope}%
\begin{pgfscope}%
\pgfpathrectangle{\pgfqpoint{0.634445in}{0.372083in}}{\pgfqpoint{5.615555in}{4.277917in}}%
\pgfusepath{clip}%
\pgfsetbuttcap%
\pgfsetmiterjoin%
\definecolor{currentfill}{rgb}{1.000000,0.498039,0.054902}%
\pgfsetfillcolor{currentfill}%
\pgfsetlinewidth{0.000000pt}%
\definecolor{currentstroke}{rgb}{0.000000,0.000000,0.000000}%
\pgfsetstrokecolor{currentstroke}%
\pgfsetstrokeopacity{0.000000}%
\pgfsetdash{}{0pt}%
\pgfpathmoveto{\pgfqpoint{2.529695in}{0.372083in}}%
\pgfpathlineto{\pgfqpoint{2.670084in}{0.372083in}}%
\pgfpathlineto{\pgfqpoint{2.670084in}{1.698932in}}%
\pgfpathlineto{\pgfqpoint{2.529695in}{1.698932in}}%
\pgfpathlineto{\pgfqpoint{2.529695in}{0.372083in}}%
\pgfpathclose%
\pgfusepath{fill}%
\end{pgfscope}%
\begin{pgfscope}%
\pgfpathrectangle{\pgfqpoint{0.634445in}{0.372083in}}{\pgfqpoint{5.615555in}{4.277917in}}%
\pgfusepath{clip}%
\pgfsetbuttcap%
\pgfsetmiterjoin%
\definecolor{currentfill}{rgb}{1.000000,0.498039,0.054902}%
\pgfsetfillcolor{currentfill}%
\pgfsetlinewidth{0.000000pt}%
\definecolor{currentstroke}{rgb}{0.000000,0.000000,0.000000}%
\pgfsetstrokecolor{currentstroke}%
\pgfsetstrokeopacity{0.000000}%
\pgfsetdash{}{0pt}%
\pgfpathmoveto{\pgfqpoint{3.933584in}{0.372083in}}%
\pgfpathlineto{\pgfqpoint{4.073972in}{0.372083in}}%
\pgfpathlineto{\pgfqpoint{4.073972in}{0.684283in}}%
\pgfpathlineto{\pgfqpoint{3.933584in}{0.684283in}}%
\pgfpathlineto{\pgfqpoint{3.933584in}{0.372083in}}%
\pgfpathclose%
\pgfusepath{fill}%
\end{pgfscope}%
\begin{pgfscope}%
\pgfpathrectangle{\pgfqpoint{0.634445in}{0.372083in}}{\pgfqpoint{5.615555in}{4.277917in}}%
\pgfusepath{clip}%
\pgfsetbuttcap%
\pgfsetmiterjoin%
\definecolor{currentfill}{rgb}{1.000000,0.498039,0.054902}%
\pgfsetfillcolor{currentfill}%
\pgfsetlinewidth{0.000000pt}%
\definecolor{currentstroke}{rgb}{0.000000,0.000000,0.000000}%
\pgfsetstrokecolor{currentstroke}%
\pgfsetstrokeopacity{0.000000}%
\pgfsetdash{}{0pt}%
\pgfpathmoveto{\pgfqpoint{5.337472in}{0.372083in}}%
\pgfpathlineto{\pgfqpoint{5.477861in}{0.372083in}}%
\pgfpathlineto{\pgfqpoint{5.477861in}{0.590623in}}%
\pgfpathlineto{\pgfqpoint{5.337472in}{0.590623in}}%
\pgfpathlineto{\pgfqpoint{5.337472in}{0.372083in}}%
\pgfpathclose%
\pgfusepath{fill}%
\end{pgfscope}%
\begin{pgfscope}%
\pgfpathrectangle{\pgfqpoint{0.634445in}{0.372083in}}{\pgfqpoint{5.615555in}{4.277917in}}%
\pgfusepath{clip}%
\pgfsetbuttcap%
\pgfsetmiterjoin%
\definecolor{currentfill}{rgb}{0.172549,0.627451,0.172549}%
\pgfsetfillcolor{currentfill}%
\pgfsetlinewidth{0.000000pt}%
\definecolor{currentstroke}{rgb}{0.000000,0.000000,0.000000}%
\pgfsetstrokecolor{currentstroke}%
\pgfsetstrokeopacity{0.000000}%
\pgfsetdash{}{0pt}%
\pgfpathmoveto{\pgfqpoint{1.266195in}{0.372083in}}%
\pgfpathlineto{\pgfqpoint{1.406584in}{0.372083in}}%
\pgfpathlineto{\pgfqpoint{1.406584in}{3.197491in}}%
\pgfpathlineto{\pgfqpoint{1.266195in}{3.197491in}}%
\pgfpathlineto{\pgfqpoint{1.266195in}{0.372083in}}%
\pgfpathclose%
\pgfusepath{fill}%
\end{pgfscope}%
\begin{pgfscope}%
\pgfpathrectangle{\pgfqpoint{0.634445in}{0.372083in}}{\pgfqpoint{5.615555in}{4.277917in}}%
\pgfusepath{clip}%
\pgfsetbuttcap%
\pgfsetmiterjoin%
\definecolor{currentfill}{rgb}{0.172549,0.627451,0.172549}%
\pgfsetfillcolor{currentfill}%
\pgfsetlinewidth{0.000000pt}%
\definecolor{currentstroke}{rgb}{0.000000,0.000000,0.000000}%
\pgfsetstrokecolor{currentstroke}%
\pgfsetstrokeopacity{0.000000}%
\pgfsetdash{}{0pt}%
\pgfpathmoveto{\pgfqpoint{2.670084in}{0.372083in}}%
\pgfpathlineto{\pgfqpoint{2.810473in}{0.372083in}}%
\pgfpathlineto{\pgfqpoint{2.810473in}{1.589662in}}%
\pgfpathlineto{\pgfqpoint{2.670084in}{1.589662in}}%
\pgfpathlineto{\pgfqpoint{2.670084in}{0.372083in}}%
\pgfpathclose%
\pgfusepath{fill}%
\end{pgfscope}%
\begin{pgfscope}%
\pgfpathrectangle{\pgfqpoint{0.634445in}{0.372083in}}{\pgfqpoint{5.615555in}{4.277917in}}%
\pgfusepath{clip}%
\pgfsetbuttcap%
\pgfsetmiterjoin%
\definecolor{currentfill}{rgb}{0.172549,0.627451,0.172549}%
\pgfsetfillcolor{currentfill}%
\pgfsetlinewidth{0.000000pt}%
\definecolor{currentstroke}{rgb}{0.000000,0.000000,0.000000}%
\pgfsetstrokecolor{currentstroke}%
\pgfsetstrokeopacity{0.000000}%
\pgfsetdash{}{0pt}%
\pgfpathmoveto{\pgfqpoint{4.073972in}{0.372083in}}%
\pgfpathlineto{\pgfqpoint{4.214361in}{0.372083in}}%
\pgfpathlineto{\pgfqpoint{4.214361in}{0.668673in}}%
\pgfpathlineto{\pgfqpoint{4.073972in}{0.668673in}}%
\pgfpathlineto{\pgfqpoint{4.073972in}{0.372083in}}%
\pgfpathclose%
\pgfusepath{fill}%
\end{pgfscope}%
\begin{pgfscope}%
\pgfpathrectangle{\pgfqpoint{0.634445in}{0.372083in}}{\pgfqpoint{5.615555in}{4.277917in}}%
\pgfusepath{clip}%
\pgfsetbuttcap%
\pgfsetmiterjoin%
\definecolor{currentfill}{rgb}{0.172549,0.627451,0.172549}%
\pgfsetfillcolor{currentfill}%
\pgfsetlinewidth{0.000000pt}%
\definecolor{currentstroke}{rgb}{0.000000,0.000000,0.000000}%
\pgfsetstrokecolor{currentstroke}%
\pgfsetstrokeopacity{0.000000}%
\pgfsetdash{}{0pt}%
\pgfpathmoveto{\pgfqpoint{5.477861in}{0.372083in}}%
\pgfpathlineto{\pgfqpoint{5.618250in}{0.372083in}}%
\pgfpathlineto{\pgfqpoint{5.618250in}{0.668673in}}%
\pgfpathlineto{\pgfqpoint{5.477861in}{0.668673in}}%
\pgfpathlineto{\pgfqpoint{5.477861in}{0.372083in}}%
\pgfpathclose%
\pgfusepath{fill}%
\end{pgfscope}%
\begin{pgfscope}%
\pgfpathrectangle{\pgfqpoint{0.634445in}{0.372083in}}{\pgfqpoint{5.615555in}{4.277917in}}%
\pgfusepath{clip}%
\pgfsetbuttcap%
\pgfsetmiterjoin%
\definecolor{currentfill}{rgb}{0.839216,0.152941,0.156863}%
\pgfsetfillcolor{currentfill}%
\pgfsetlinewidth{0.000000pt}%
\definecolor{currentstroke}{rgb}{0.000000,0.000000,0.000000}%
\pgfsetstrokecolor{currentstroke}%
\pgfsetstrokeopacity{0.000000}%
\pgfsetdash{}{0pt}%
\pgfpathmoveto{\pgfqpoint{1.406584in}{0.372083in}}%
\pgfpathlineto{\pgfqpoint{1.546973in}{0.372083in}}%
\pgfpathlineto{\pgfqpoint{1.546973in}{2.276502in}}%
\pgfpathlineto{\pgfqpoint{1.406584in}{2.276502in}}%
\pgfpathlineto{\pgfqpoint{1.406584in}{0.372083in}}%
\pgfpathclose%
\pgfusepath{fill}%
\end{pgfscope}%
\begin{pgfscope}%
\pgfpathrectangle{\pgfqpoint{0.634445in}{0.372083in}}{\pgfqpoint{5.615555in}{4.277917in}}%
\pgfusepath{clip}%
\pgfsetbuttcap%
\pgfsetmiterjoin%
\definecolor{currentfill}{rgb}{0.839216,0.152941,0.156863}%
\pgfsetfillcolor{currentfill}%
\pgfsetlinewidth{0.000000pt}%
\definecolor{currentstroke}{rgb}{0.000000,0.000000,0.000000}%
\pgfsetstrokecolor{currentstroke}%
\pgfsetstrokeopacity{0.000000}%
\pgfsetdash{}{0pt}%
\pgfpathmoveto{\pgfqpoint{2.810473in}{0.372083in}}%
\pgfpathlineto{\pgfqpoint{2.950861in}{0.372083in}}%
\pgfpathlineto{\pgfqpoint{2.950861in}{1.855032in}}%
\pgfpathlineto{\pgfqpoint{2.810473in}{1.855032in}}%
\pgfpathlineto{\pgfqpoint{2.810473in}{0.372083in}}%
\pgfpathclose%
\pgfusepath{fill}%
\end{pgfscope}%
\begin{pgfscope}%
\pgfpathrectangle{\pgfqpoint{0.634445in}{0.372083in}}{\pgfqpoint{5.615555in}{4.277917in}}%
\pgfusepath{clip}%
\pgfsetbuttcap%
\pgfsetmiterjoin%
\definecolor{currentfill}{rgb}{0.839216,0.152941,0.156863}%
\pgfsetfillcolor{currentfill}%
\pgfsetlinewidth{0.000000pt}%
\definecolor{currentstroke}{rgb}{0.000000,0.000000,0.000000}%
\pgfsetstrokecolor{currentstroke}%
\pgfsetstrokeopacity{0.000000}%
\pgfsetdash{}{0pt}%
\pgfpathmoveto{\pgfqpoint{4.214361in}{0.372083in}}%
\pgfpathlineto{\pgfqpoint{4.354750in}{0.372083in}}%
\pgfpathlineto{\pgfqpoint{4.354750in}{1.620882in}}%
\pgfpathlineto{\pgfqpoint{4.214361in}{1.620882in}}%
\pgfpathlineto{\pgfqpoint{4.214361in}{0.372083in}}%
\pgfpathclose%
\pgfusepath{fill}%
\end{pgfscope}%
\begin{pgfscope}%
\pgfpathrectangle{\pgfqpoint{0.634445in}{0.372083in}}{\pgfqpoint{5.615555in}{4.277917in}}%
\pgfusepath{clip}%
\pgfsetbuttcap%
\pgfsetmiterjoin%
\definecolor{currentfill}{rgb}{0.839216,0.152941,0.156863}%
\pgfsetfillcolor{currentfill}%
\pgfsetlinewidth{0.000000pt}%
\definecolor{currentstroke}{rgb}{0.000000,0.000000,0.000000}%
\pgfsetstrokecolor{currentstroke}%
\pgfsetstrokeopacity{0.000000}%
\pgfsetdash{}{0pt}%
\pgfpathmoveto{\pgfqpoint{5.618250in}{0.372083in}}%
\pgfpathlineto{\pgfqpoint{5.758639in}{0.372083in}}%
\pgfpathlineto{\pgfqpoint{5.758639in}{0.372083in}}%
\pgfpathlineto{\pgfqpoint{5.618250in}{0.372083in}}%
\pgfpathlineto{\pgfqpoint{5.618250in}{0.372083in}}%
\pgfpathclose%
\pgfusepath{fill}%
\end{pgfscope}%
\begin{pgfscope}%
\pgfpathrectangle{\pgfqpoint{0.634445in}{0.372083in}}{\pgfqpoint{5.615555in}{4.277917in}}%
\pgfusepath{clip}%
\pgfsetbuttcap%
\pgfsetmiterjoin%
\definecolor{currentfill}{rgb}{0.580392,0.403922,0.741176}%
\pgfsetfillcolor{currentfill}%
\pgfsetlinewidth{0.000000pt}%
\definecolor{currentstroke}{rgb}{0.000000,0.000000,0.000000}%
\pgfsetstrokecolor{currentstroke}%
\pgfsetstrokeopacity{0.000000}%
\pgfsetdash{}{0pt}%
\pgfpathmoveto{\pgfqpoint{1.546973in}{0.372083in}}%
\pgfpathlineto{\pgfqpoint{1.687362in}{0.372083in}}%
\pgfpathlineto{\pgfqpoint{1.687362in}{4.446290in}}%
\pgfpathlineto{\pgfqpoint{1.546973in}{4.446290in}}%
\pgfpathlineto{\pgfqpoint{1.546973in}{0.372083in}}%
\pgfpathclose%
\pgfusepath{fill}%
\end{pgfscope}%
\begin{pgfscope}%
\pgfpathrectangle{\pgfqpoint{0.634445in}{0.372083in}}{\pgfqpoint{5.615555in}{4.277917in}}%
\pgfusepath{clip}%
\pgfsetbuttcap%
\pgfsetmiterjoin%
\definecolor{currentfill}{rgb}{0.580392,0.403922,0.741176}%
\pgfsetfillcolor{currentfill}%
\pgfsetlinewidth{0.000000pt}%
\definecolor{currentstroke}{rgb}{0.000000,0.000000,0.000000}%
\pgfsetstrokecolor{currentstroke}%
\pgfsetstrokeopacity{0.000000}%
\pgfsetdash{}{0pt}%
\pgfpathmoveto{\pgfqpoint{2.950861in}{0.372083in}}%
\pgfpathlineto{\pgfqpoint{3.091250in}{0.372083in}}%
\pgfpathlineto{\pgfqpoint{3.091250in}{0.731113in}}%
\pgfpathlineto{\pgfqpoint{2.950861in}{0.731113in}}%
\pgfpathlineto{\pgfqpoint{2.950861in}{0.372083in}}%
\pgfpathclose%
\pgfusepath{fill}%
\end{pgfscope}%
\begin{pgfscope}%
\pgfpathrectangle{\pgfqpoint{0.634445in}{0.372083in}}{\pgfqpoint{5.615555in}{4.277917in}}%
\pgfusepath{clip}%
\pgfsetbuttcap%
\pgfsetmiterjoin%
\definecolor{currentfill}{rgb}{0.580392,0.403922,0.741176}%
\pgfsetfillcolor{currentfill}%
\pgfsetlinewidth{0.000000pt}%
\definecolor{currentstroke}{rgb}{0.000000,0.000000,0.000000}%
\pgfsetstrokecolor{currentstroke}%
\pgfsetstrokeopacity{0.000000}%
\pgfsetdash{}{0pt}%
\pgfpathmoveto{\pgfqpoint{4.354750in}{0.372083in}}%
\pgfpathlineto{\pgfqpoint{4.495139in}{0.372083in}}%
\pgfpathlineto{\pgfqpoint{4.495139in}{0.575013in}}%
\pgfpathlineto{\pgfqpoint{4.354750in}{0.575013in}}%
\pgfpathlineto{\pgfqpoint{4.354750in}{0.372083in}}%
\pgfpathclose%
\pgfusepath{fill}%
\end{pgfscope}%
\begin{pgfscope}%
\pgfpathrectangle{\pgfqpoint{0.634445in}{0.372083in}}{\pgfqpoint{5.615555in}{4.277917in}}%
\pgfusepath{clip}%
\pgfsetbuttcap%
\pgfsetmiterjoin%
\definecolor{currentfill}{rgb}{0.580392,0.403922,0.741176}%
\pgfsetfillcolor{currentfill}%
\pgfsetlinewidth{0.000000pt}%
\definecolor{currentstroke}{rgb}{0.000000,0.000000,0.000000}%
\pgfsetstrokecolor{currentstroke}%
\pgfsetstrokeopacity{0.000000}%
\pgfsetdash{}{0pt}%
\pgfpathmoveto{\pgfqpoint{5.758639in}{0.372083in}}%
\pgfpathlineto{\pgfqpoint{5.899028in}{0.372083in}}%
\pgfpathlineto{\pgfqpoint{5.899028in}{0.372083in}}%
\pgfpathlineto{\pgfqpoint{5.758639in}{0.372083in}}%
\pgfpathlineto{\pgfqpoint{5.758639in}{0.372083in}}%
\pgfpathclose%
\pgfusepath{fill}%
\end{pgfscope}%
\begin{pgfscope}%
\pgfsetbuttcap%
\pgfsetroundjoin%
\definecolor{currentfill}{rgb}{0.000000,0.000000,0.000000}%
\pgfsetfillcolor{currentfill}%
\pgfsetlinewidth{0.803000pt}%
\definecolor{currentstroke}{rgb}{0.000000,0.000000,0.000000}%
\pgfsetstrokecolor{currentstroke}%
\pgfsetdash{}{0pt}%
\pgfsys@defobject{currentmarker}{\pgfqpoint{0.000000in}{-0.048611in}}{\pgfqpoint{0.000000in}{0.000000in}}{%
\pgfpathmoveto{\pgfqpoint{0.000000in}{0.000000in}}%
\pgfpathlineto{\pgfqpoint{0.000000in}{-0.048611in}}%
\pgfusepath{stroke,fill}%
}%
\begin{pgfscope}%
\pgfsys@transformshift{1.336389in}{0.372083in}%
\pgfsys@useobject{currentmarker}{}%
\end{pgfscope}%
\end{pgfscope}%
\begin{pgfscope}%
\definecolor{textcolor}{rgb}{0.000000,0.000000,0.000000}%
\pgfsetstrokecolor{textcolor}%
\pgfsetfillcolor{textcolor}%
\pgftext[x=1.336389in,y=0.274861in,,top]{\color{textcolor}\rmfamily\fontsize{10.000000}{12.000000}\selectfont Easier to read}%
\end{pgfscope}%
\begin{pgfscope}%
\pgfsetbuttcap%
\pgfsetroundjoin%
\definecolor{currentfill}{rgb}{0.000000,0.000000,0.000000}%
\pgfsetfillcolor{currentfill}%
\pgfsetlinewidth{0.803000pt}%
\definecolor{currentstroke}{rgb}{0.000000,0.000000,0.000000}%
\pgfsetstrokecolor{currentstroke}%
\pgfsetdash{}{0pt}%
\pgfsys@defobject{currentmarker}{\pgfqpoint{0.000000in}{-0.048611in}}{\pgfqpoint{0.000000in}{0.000000in}}{%
\pgfpathmoveto{\pgfqpoint{0.000000in}{0.000000in}}%
\pgfpathlineto{\pgfqpoint{0.000000in}{-0.048611in}}%
\pgfusepath{stroke,fill}%
}%
\begin{pgfscope}%
\pgfsys@transformshift{2.740278in}{0.372083in}%
\pgfsys@useobject{currentmarker}{}%
\end{pgfscope}%
\end{pgfscope}%
\begin{pgfscope}%
\definecolor{textcolor}{rgb}{0.000000,0.000000,0.000000}%
\pgfsetstrokecolor{textcolor}%
\pgfsetfillcolor{textcolor}%
\pgftext[x=2.740278in,y=0.274861in,,top]{\color{textcolor}\rmfamily\fontsize{10.000000}{12.000000}\selectfont Equal to read}%
\end{pgfscope}%
\begin{pgfscope}%
\pgfsetbuttcap%
\pgfsetroundjoin%
\definecolor{currentfill}{rgb}{0.000000,0.000000,0.000000}%
\pgfsetfillcolor{currentfill}%
\pgfsetlinewidth{0.803000pt}%
\definecolor{currentstroke}{rgb}{0.000000,0.000000,0.000000}%
\pgfsetstrokecolor{currentstroke}%
\pgfsetdash{}{0pt}%
\pgfsys@defobject{currentmarker}{\pgfqpoint{0.000000in}{-0.048611in}}{\pgfqpoint{0.000000in}{0.000000in}}{%
\pgfpathmoveto{\pgfqpoint{0.000000in}{0.000000in}}%
\pgfpathlineto{\pgfqpoint{0.000000in}{-0.048611in}}%
\pgfusepath{stroke,fill}%
}%
\begin{pgfscope}%
\pgfsys@transformshift{4.144167in}{0.372083in}%
\pgfsys@useobject{currentmarker}{}%
\end{pgfscope}%
\end{pgfscope}%
\begin{pgfscope}%
\definecolor{textcolor}{rgb}{0.000000,0.000000,0.000000}%
\pgfsetstrokecolor{textcolor}%
\pgfsetfillcolor{textcolor}%
\pgftext[x=4.144167in,y=0.274861in,,top]{\color{textcolor}\rmfamily\fontsize{10.000000}{12.000000}\selectfont More difficult}%
\end{pgfscope}%
\begin{pgfscope}%
\pgfsetbuttcap%
\pgfsetroundjoin%
\definecolor{currentfill}{rgb}{0.000000,0.000000,0.000000}%
\pgfsetfillcolor{currentfill}%
\pgfsetlinewidth{0.803000pt}%
\definecolor{currentstroke}{rgb}{0.000000,0.000000,0.000000}%
\pgfsetstrokecolor{currentstroke}%
\pgfsetdash{}{0pt}%
\pgfsys@defobject{currentmarker}{\pgfqpoint{0.000000in}{-0.048611in}}{\pgfqpoint{0.000000in}{0.000000in}}{%
\pgfpathmoveto{\pgfqpoint{0.000000in}{0.000000in}}%
\pgfpathlineto{\pgfqpoint{0.000000in}{-0.048611in}}%
\pgfusepath{stroke,fill}%
}%
\begin{pgfscope}%
\pgfsys@transformshift{5.548056in}{0.372083in}%
\pgfsys@useobject{currentmarker}{}%
\end{pgfscope}%
\end{pgfscope}%
\begin{pgfscope}%
\definecolor{textcolor}{rgb}{0.000000,0.000000,0.000000}%
\pgfsetstrokecolor{textcolor}%
\pgfsetfillcolor{textcolor}%
\pgftext[x=5.548056in,y=0.274861in,,top]{\color{textcolor}\rmfamily\fontsize{10.000000}{12.000000}\selectfont No Simplification}%
\end{pgfscope}%
\begin{pgfscope}%
\pgfsetbuttcap%
\pgfsetroundjoin%
\definecolor{currentfill}{rgb}{0.000000,0.000000,0.000000}%
\pgfsetfillcolor{currentfill}%
\pgfsetlinewidth{0.803000pt}%
\definecolor{currentstroke}{rgb}{0.000000,0.000000,0.000000}%
\pgfsetstrokecolor{currentstroke}%
\pgfsetdash{}{0pt}%
\pgfsys@defobject{currentmarker}{\pgfqpoint{-0.048611in}{0.000000in}}{\pgfqpoint{-0.000000in}{0.000000in}}{%
\pgfpathmoveto{\pgfqpoint{-0.000000in}{0.000000in}}%
\pgfpathlineto{\pgfqpoint{-0.048611in}{0.000000in}}%
\pgfusepath{stroke,fill}%
}%
\begin{pgfscope}%
\pgfsys@transformshift{0.634445in}{0.372083in}%
\pgfsys@useobject{currentmarker}{}%
\end{pgfscope}%
\end{pgfscope}%
\begin{pgfscope}%
\definecolor{textcolor}{rgb}{0.000000,0.000000,0.000000}%
\pgfsetstrokecolor{textcolor}%
\pgfsetfillcolor{textcolor}%
\pgftext[x=0.467778in, y=0.323889in, left, base]{\color{textcolor}\rmfamily\fontsize{10.000000}{12.000000}\selectfont \(\displaystyle {0}\)}%
\end{pgfscope}%
\begin{pgfscope}%
\pgfsetbuttcap%
\pgfsetroundjoin%
\definecolor{currentfill}{rgb}{0.000000,0.000000,0.000000}%
\pgfsetfillcolor{currentfill}%
\pgfsetlinewidth{0.803000pt}%
\definecolor{currentstroke}{rgb}{0.000000,0.000000,0.000000}%
\pgfsetstrokecolor{currentstroke}%
\pgfsetdash{}{0pt}%
\pgfsys@defobject{currentmarker}{\pgfqpoint{-0.048611in}{0.000000in}}{\pgfqpoint{-0.000000in}{0.000000in}}{%
\pgfpathmoveto{\pgfqpoint{-0.000000in}{0.000000in}}%
\pgfpathlineto{\pgfqpoint{-0.048611in}{0.000000in}}%
\pgfusepath{stroke,fill}%
}%
\begin{pgfscope}%
\pgfsys@transformshift{0.634445in}{1.152583in}%
\pgfsys@useobject{currentmarker}{}%
\end{pgfscope}%
\end{pgfscope}%
\begin{pgfscope}%
\definecolor{textcolor}{rgb}{0.000000,0.000000,0.000000}%
\pgfsetstrokecolor{textcolor}%
\pgfsetfillcolor{textcolor}%
\pgftext[x=0.398333in, y=1.104388in, left, base]{\color{textcolor}\rmfamily\fontsize{10.000000}{12.000000}\selectfont \(\displaystyle {50}\)}%
\end{pgfscope}%
\begin{pgfscope}%
\pgfsetbuttcap%
\pgfsetroundjoin%
\definecolor{currentfill}{rgb}{0.000000,0.000000,0.000000}%
\pgfsetfillcolor{currentfill}%
\pgfsetlinewidth{0.803000pt}%
\definecolor{currentstroke}{rgb}{0.000000,0.000000,0.000000}%
\pgfsetstrokecolor{currentstroke}%
\pgfsetdash{}{0pt}%
\pgfsys@defobject{currentmarker}{\pgfqpoint{-0.048611in}{0.000000in}}{\pgfqpoint{-0.000000in}{0.000000in}}{%
\pgfpathmoveto{\pgfqpoint{-0.000000in}{0.000000in}}%
\pgfpathlineto{\pgfqpoint{-0.048611in}{0.000000in}}%
\pgfusepath{stroke,fill}%
}%
\begin{pgfscope}%
\pgfsys@transformshift{0.634445in}{1.933082in}%
\pgfsys@useobject{currentmarker}{}%
\end{pgfscope}%
\end{pgfscope}%
\begin{pgfscope}%
\definecolor{textcolor}{rgb}{0.000000,0.000000,0.000000}%
\pgfsetstrokecolor{textcolor}%
\pgfsetfillcolor{textcolor}%
\pgftext[x=0.328889in, y=1.884887in, left, base]{\color{textcolor}\rmfamily\fontsize{10.000000}{12.000000}\selectfont \(\displaystyle {100}\)}%
\end{pgfscope}%
\begin{pgfscope}%
\pgfsetbuttcap%
\pgfsetroundjoin%
\definecolor{currentfill}{rgb}{0.000000,0.000000,0.000000}%
\pgfsetfillcolor{currentfill}%
\pgfsetlinewidth{0.803000pt}%
\definecolor{currentstroke}{rgb}{0.000000,0.000000,0.000000}%
\pgfsetstrokecolor{currentstroke}%
\pgfsetdash{}{0pt}%
\pgfsys@defobject{currentmarker}{\pgfqpoint{-0.048611in}{0.000000in}}{\pgfqpoint{-0.000000in}{0.000000in}}{%
\pgfpathmoveto{\pgfqpoint{-0.000000in}{0.000000in}}%
\pgfpathlineto{\pgfqpoint{-0.048611in}{0.000000in}}%
\pgfusepath{stroke,fill}%
}%
\begin{pgfscope}%
\pgfsys@transformshift{0.634445in}{2.713581in}%
\pgfsys@useobject{currentmarker}{}%
\end{pgfscope}%
\end{pgfscope}%
\begin{pgfscope}%
\definecolor{textcolor}{rgb}{0.000000,0.000000,0.000000}%
\pgfsetstrokecolor{textcolor}%
\pgfsetfillcolor{textcolor}%
\pgftext[x=0.328889in, y=2.665387in, left, base]{\color{textcolor}\rmfamily\fontsize{10.000000}{12.000000}\selectfont \(\displaystyle {150}\)}%
\end{pgfscope}%
\begin{pgfscope}%
\pgfsetbuttcap%
\pgfsetroundjoin%
\definecolor{currentfill}{rgb}{0.000000,0.000000,0.000000}%
\pgfsetfillcolor{currentfill}%
\pgfsetlinewidth{0.803000pt}%
\definecolor{currentstroke}{rgb}{0.000000,0.000000,0.000000}%
\pgfsetstrokecolor{currentstroke}%
\pgfsetdash{}{0pt}%
\pgfsys@defobject{currentmarker}{\pgfqpoint{-0.048611in}{0.000000in}}{\pgfqpoint{-0.000000in}{0.000000in}}{%
\pgfpathmoveto{\pgfqpoint{-0.000000in}{0.000000in}}%
\pgfpathlineto{\pgfqpoint{-0.048611in}{0.000000in}}%
\pgfusepath{stroke,fill}%
}%
\begin{pgfscope}%
\pgfsys@transformshift{0.634445in}{3.494080in}%
\pgfsys@useobject{currentmarker}{}%
\end{pgfscope}%
\end{pgfscope}%
\begin{pgfscope}%
\definecolor{textcolor}{rgb}{0.000000,0.000000,0.000000}%
\pgfsetstrokecolor{textcolor}%
\pgfsetfillcolor{textcolor}%
\pgftext[x=0.328889in, y=3.445886in, left, base]{\color{textcolor}\rmfamily\fontsize{10.000000}{12.000000}\selectfont \(\displaystyle {200}\)}%
\end{pgfscope}%
\begin{pgfscope}%
\pgfsetbuttcap%
\pgfsetroundjoin%
\definecolor{currentfill}{rgb}{0.000000,0.000000,0.000000}%
\pgfsetfillcolor{currentfill}%
\pgfsetlinewidth{0.803000pt}%
\definecolor{currentstroke}{rgb}{0.000000,0.000000,0.000000}%
\pgfsetstrokecolor{currentstroke}%
\pgfsetdash{}{0pt}%
\pgfsys@defobject{currentmarker}{\pgfqpoint{-0.048611in}{0.000000in}}{\pgfqpoint{-0.000000in}{0.000000in}}{%
\pgfpathmoveto{\pgfqpoint{-0.000000in}{0.000000in}}%
\pgfpathlineto{\pgfqpoint{-0.048611in}{0.000000in}}%
\pgfusepath{stroke,fill}%
}%
\begin{pgfscope}%
\pgfsys@transformshift{0.634445in}{4.274580in}%
\pgfsys@useobject{currentmarker}{}%
\end{pgfscope}%
\end{pgfscope}%
\begin{pgfscope}%
\definecolor{textcolor}{rgb}{0.000000,0.000000,0.000000}%
\pgfsetstrokecolor{textcolor}%
\pgfsetfillcolor{textcolor}%
\pgftext[x=0.328889in, y=4.226385in, left, base]{\color{textcolor}\rmfamily\fontsize{10.000000}{12.000000}\selectfont \(\displaystyle {250}\)}%
\end{pgfscope}%
\begin{pgfscope}%
\definecolor{textcolor}{rgb}{0.000000,0.000000,0.000000}%
\pgfsetstrokecolor{textcolor}%
\pgfsetfillcolor{textcolor}%
\pgftext[x=0.273333in,y=2.511042in,,bottom,rotate=90.000000]{\color{textcolor}\rmfamily\fontsize{10.000000}{12.000000}\selectfont Instances}%
\end{pgfscope}%
\begin{pgfscope}%
\pgfsetrectcap%
\pgfsetmiterjoin%
\pgfsetlinewidth{0.803000pt}%
\definecolor{currentstroke}{rgb}{0.000000,0.000000,0.000000}%
\pgfsetstrokecolor{currentstroke}%
\pgfsetdash{}{0pt}%
\pgfpathmoveto{\pgfqpoint{0.634445in}{0.372083in}}%
\pgfpathlineto{\pgfqpoint{0.634445in}{4.650000in}}%
\pgfusepath{stroke}%
\end{pgfscope}%
\begin{pgfscope}%
\pgfsetrectcap%
\pgfsetmiterjoin%
\pgfsetlinewidth{0.803000pt}%
\definecolor{currentstroke}{rgb}{0.000000,0.000000,0.000000}%
\pgfsetstrokecolor{currentstroke}%
\pgfsetdash{}{0pt}%
\pgfpathmoveto{\pgfqpoint{6.250000in}{0.372083in}}%
\pgfpathlineto{\pgfqpoint{6.250000in}{4.650000in}}%
\pgfusepath{stroke}%
\end{pgfscope}%
\begin{pgfscope}%
\pgfsetrectcap%
\pgfsetmiterjoin%
\pgfsetlinewidth{0.803000pt}%
\definecolor{currentstroke}{rgb}{0.000000,0.000000,0.000000}%
\pgfsetstrokecolor{currentstroke}%
\pgfsetdash{}{0pt}%
\pgfpathmoveto{\pgfqpoint{0.634445in}{0.372083in}}%
\pgfpathlineto{\pgfqpoint{6.250000in}{0.372083in}}%
\pgfusepath{stroke}%
\end{pgfscope}%
\begin{pgfscope}%
\pgfsetrectcap%
\pgfsetmiterjoin%
\pgfsetlinewidth{0.803000pt}%
\definecolor{currentstroke}{rgb}{0.000000,0.000000,0.000000}%
\pgfsetstrokecolor{currentstroke}%
\pgfsetdash{}{0pt}%
\pgfpathmoveto{\pgfqpoint{0.634445in}{4.650000in}}%
\pgfpathlineto{\pgfqpoint{6.250000in}{4.650000in}}%
\pgfusepath{stroke}%
\end{pgfscope}%
\begin{pgfscope}%
\pgfsetbuttcap%
\pgfsetmiterjoin%
\definecolor{currentfill}{rgb}{1.000000,1.000000,1.000000}%
\pgfsetfillcolor{currentfill}%
\pgfsetfillopacity{0.800000}%
\pgfsetlinewidth{1.003750pt}%
\definecolor{currentstroke}{rgb}{0.800000,0.800000,0.800000}%
\pgfsetstrokecolor{currentstroke}%
\pgfsetstrokeopacity{0.800000}%
\pgfsetdash{}{0pt}%
\pgfpathmoveto{\pgfqpoint{4.926528in}{3.570834in}}%
\pgfpathlineto{\pgfqpoint{6.152778in}{3.570834in}}%
\pgfpathquadraticcurveto{\pgfqpoint{6.180556in}{3.570834in}}{\pgfqpoint{6.180556in}{3.598612in}}%
\pgfpathlineto{\pgfqpoint{6.180556in}{4.552778in}}%
\pgfpathquadraticcurveto{\pgfqpoint{6.180556in}{4.580556in}}{\pgfqpoint{6.152778in}{4.580556in}}%
\pgfpathlineto{\pgfqpoint{4.926528in}{4.580556in}}%
\pgfpathquadraticcurveto{\pgfqpoint{4.898750in}{4.580556in}}{\pgfqpoint{4.898750in}{4.552778in}}%
\pgfpathlineto{\pgfqpoint{4.898750in}{3.598612in}}%
\pgfpathquadraticcurveto{\pgfqpoint{4.898750in}{3.570834in}}{\pgfqpoint{4.926528in}{3.570834in}}%
\pgfpathlineto{\pgfqpoint{4.926528in}{3.570834in}}%
\pgfpathclose%
\pgfusepath{stroke,fill}%
\end{pgfscope}%
\begin{pgfscope}%
\pgfsetbuttcap%
\pgfsetmiterjoin%
\definecolor{currentfill}{rgb}{0.121569,0.466667,0.705882}%
\pgfsetfillcolor{currentfill}%
\pgfsetlinewidth{0.000000pt}%
\definecolor{currentstroke}{rgb}{0.000000,0.000000,0.000000}%
\pgfsetstrokecolor{currentstroke}%
\pgfsetstrokeopacity{0.000000}%
\pgfsetdash{}{0pt}%
\pgfpathmoveto{\pgfqpoint{4.954305in}{4.427778in}}%
\pgfpathlineto{\pgfqpoint{5.232083in}{4.427778in}}%
\pgfpathlineto{\pgfqpoint{5.232083in}{4.525000in}}%
\pgfpathlineto{\pgfqpoint{4.954305in}{4.525000in}}%
\pgfpathlineto{\pgfqpoint{4.954305in}{4.427778in}}%
\pgfpathclose%
\pgfusepath{fill}%
\end{pgfscope}%
\begin{pgfscope}%
\definecolor{textcolor}{rgb}{0.000000,0.000000,0.000000}%
\pgfsetstrokecolor{textcolor}%
\pgfsetfillcolor{textcolor}%
\pgftext[x=5.343194in,y=4.427778in,left,base]{\color{textcolor}\rmfamily\fontsize{10.000000}{12.000000}\selectfont Annotator A}%
\end{pgfscope}%
\begin{pgfscope}%
\pgfsetbuttcap%
\pgfsetmiterjoin%
\definecolor{currentfill}{rgb}{1.000000,0.498039,0.054902}%
\pgfsetfillcolor{currentfill}%
\pgfsetlinewidth{0.000000pt}%
\definecolor{currentstroke}{rgb}{0.000000,0.000000,0.000000}%
\pgfsetstrokecolor{currentstroke}%
\pgfsetstrokeopacity{0.000000}%
\pgfsetdash{}{0pt}%
\pgfpathmoveto{\pgfqpoint{4.954305in}{4.234167in}}%
\pgfpathlineto{\pgfqpoint{5.232083in}{4.234167in}}%
\pgfpathlineto{\pgfqpoint{5.232083in}{4.331389in}}%
\pgfpathlineto{\pgfqpoint{4.954305in}{4.331389in}}%
\pgfpathlineto{\pgfqpoint{4.954305in}{4.234167in}}%
\pgfpathclose%
\pgfusepath{fill}%
\end{pgfscope}%
\begin{pgfscope}%
\definecolor{textcolor}{rgb}{0.000000,0.000000,0.000000}%
\pgfsetstrokecolor{textcolor}%
\pgfsetfillcolor{textcolor}%
\pgftext[x=5.343194in,y=4.234167in,left,base]{\color{textcolor}\rmfamily\fontsize{10.000000}{12.000000}\selectfont Annotator B}%
\end{pgfscope}%
\begin{pgfscope}%
\pgfsetbuttcap%
\pgfsetmiterjoin%
\definecolor{currentfill}{rgb}{0.172549,0.627451,0.172549}%
\pgfsetfillcolor{currentfill}%
\pgfsetlinewidth{0.000000pt}%
\definecolor{currentstroke}{rgb}{0.000000,0.000000,0.000000}%
\pgfsetstrokecolor{currentstroke}%
\pgfsetstrokeopacity{0.000000}%
\pgfsetdash{}{0pt}%
\pgfpathmoveto{\pgfqpoint{4.954305in}{4.040556in}}%
\pgfpathlineto{\pgfqpoint{5.232083in}{4.040556in}}%
\pgfpathlineto{\pgfqpoint{5.232083in}{4.137778in}}%
\pgfpathlineto{\pgfqpoint{4.954305in}{4.137778in}}%
\pgfpathlineto{\pgfqpoint{4.954305in}{4.040556in}}%
\pgfpathclose%
\pgfusepath{fill}%
\end{pgfscope}%
\begin{pgfscope}%
\definecolor{textcolor}{rgb}{0.000000,0.000000,0.000000}%
\pgfsetstrokecolor{textcolor}%
\pgfsetfillcolor{textcolor}%
\pgftext[x=5.343194in,y=4.040556in,left,base]{\color{textcolor}\rmfamily\fontsize{10.000000}{12.000000}\selectfont Annotator C}%
\end{pgfscope}%
\begin{pgfscope}%
\pgfsetbuttcap%
\pgfsetmiterjoin%
\definecolor{currentfill}{rgb}{0.839216,0.152941,0.156863}%
\pgfsetfillcolor{currentfill}%
\pgfsetlinewidth{0.000000pt}%
\definecolor{currentstroke}{rgb}{0.000000,0.000000,0.000000}%
\pgfsetstrokecolor{currentstroke}%
\pgfsetstrokeopacity{0.000000}%
\pgfsetdash{}{0pt}%
\pgfpathmoveto{\pgfqpoint{4.954305in}{3.846945in}}%
\pgfpathlineto{\pgfqpoint{5.232083in}{3.846945in}}%
\pgfpathlineto{\pgfqpoint{5.232083in}{3.944167in}}%
\pgfpathlineto{\pgfqpoint{4.954305in}{3.944167in}}%
\pgfpathlineto{\pgfqpoint{4.954305in}{3.846945in}}%
\pgfpathclose%
\pgfusepath{fill}%
\end{pgfscope}%
\begin{pgfscope}%
\definecolor{textcolor}{rgb}{0.000000,0.000000,0.000000}%
\pgfsetstrokecolor{textcolor}%
\pgfsetfillcolor{textcolor}%
\pgftext[x=5.343194in,y=3.846945in,left,base]{\color{textcolor}\rmfamily\fontsize{10.000000}{12.000000}\selectfont Annotator D}%
\end{pgfscope}%
\begin{pgfscope}%
\pgfsetbuttcap%
\pgfsetmiterjoin%
\definecolor{currentfill}{rgb}{0.580392,0.403922,0.741176}%
\pgfsetfillcolor{currentfill}%
\pgfsetlinewidth{0.000000pt}%
\definecolor{currentstroke}{rgb}{0.000000,0.000000,0.000000}%
\pgfsetstrokecolor{currentstroke}%
\pgfsetstrokeopacity{0.000000}%
\pgfsetdash{}{0pt}%
\pgfpathmoveto{\pgfqpoint{4.954305in}{3.653334in}}%
\pgfpathlineto{\pgfqpoint{5.232083in}{3.653334in}}%
\pgfpathlineto{\pgfqpoint{5.232083in}{3.750556in}}%
\pgfpathlineto{\pgfqpoint{4.954305in}{3.750556in}}%
\pgfpathlineto{\pgfqpoint{4.954305in}{3.653334in}}%
\pgfpathclose%
\pgfusepath{fill}%
\end{pgfscope}%
\begin{pgfscope}%
\definecolor{textcolor}{rgb}{0.000000,0.000000,0.000000}%
\pgfsetstrokecolor{textcolor}%
\pgfsetfillcolor{textcolor}%
\pgftext[x=5.343194in,y=3.653334in,left,base]{\color{textcolor}\rmfamily\fontsize{10.000000}{12.000000}\selectfont Annotator E}%
\end{pgfscope}%
\end{pgfpicture}%
\makeatother%
\endgroup%

%% file: datasheet.tex
\section{Human Evaluation Datasheet}
\label{ann:hed}

\subsection{Paper and Supplementary Resources (Questions 1.1--1.3)}\label{sec:paper-resources}

\subsubsection*{\qsecbox{Question 1.1: Link to paper reporting the evaluation experiment. If the paper reports more than one experiment, state which experiment you're completing this sheet for. Or, if applicable, enter `for preregistration.'}}
\noindent For preregistration.

\subsubsection*{\qsecbox{Question 1.2: Link to website providing resources used in the evaluation experiment (e.g.\ system outputs, evaluation tools, etc.). If there isn't one, enter `N/A'.}}
\noindent N/A.

\subsubsection*{\qsecbox{Question 1.3: Name, affiliation and email address of person completing this sheet, and of contact author if different.}}
\noindent David Beauchemin\\david.beauchemin@ift.ulaval.ca

\subsection{System (Questions 2.1--2.5)}\label{sec:system}
\subsubsection*{\qsecbox{Question 2.1: What type of input do the evaluated system(s) take? Select all that apply. If none match, select `Other' and describe.}}\label{sec:input}

\noindent\textit{Check-box options (select all that apply)}: 
\begin{enumerate}[itemsep=0cm, leftmargin=0.5cm, label={\small $\square$}]
    \item[\checkmark] \egcvalue{raw/structured data}: numerical, symbolic, and other data, possibly structured into trees, graphs, graphical models, etc. May be the input e.g.\ to Referring Expression Generation (REG), end-to-end text generation, etc. {NB}: excludes linguistic structures.
    \item \egcvalue{deep linguistic representation (DLR)}: any of a variety of deep, underspecified, semantic representations, such as abstract meaning representations \citep[AMRs;][]{banarescu-etal-2013-abstract} or discourse representation structures \citep[DRSs;][]{kamp2013discourse}.
    \item \egcvalue{shallow linguistic representation (SLR)}: any of a variety of shallow, syntactic representations, e.g.\ Universal Dependency (UD) structures; typically the input to surface realisation.
    \item \egcvalue{text: subsentential unit of text}: a unit of text shorter than a sentence, e.g.\ Referring Expressions (REs), verb phrase, text fragment of any length; includes titles/headlines.
    \item \egcvalue{text: sentence}: a single sentence (or set of sentences).
    \item[\checkmark] \egcvalue{text: multiple sentences}: a sequence of multiple sentences, without any document structure (or a set of such sequences). 
    \item \egcvalue{text: document}: a text with document structure, such as a title, paragraph breaks or sections, e.g.\ a set of news reports for summarisation.
    \item \egcvalue{text: dialogue}: a dialogue of any length, excluding a single turn which would come under one of the other text types.
    \item \egcvalue{text: other}: input is text but doesn't match any of the above \textit{text:*} categories.
    \item \egcvalue{speech}: a recording of speech.
    \item \egcvalue{visual}: an image or video.
    \item \egcvalue{multi-modal}: catch-all value for any combination of data and/or linguistic representation and/or visual data etc.
    \item \egcvalue{control feature}: a feature or parameter specifically present to control a property of the output text, e.g.\ positive stance, formality, author style.
    \item \egcvalue{no input (human generation)}: human generation\footnote{\label{human-generation}We use the term `human generation' where the items being evaluated have been created manually, rather than generated by an automatic system.}, therefore no system inputs.
    \item \egcvalue{other (please specify)}: if input is none of the above, choose this option and describe it.   
\end{enumerate}

\subsubsection*{\qsecbox{Question 2.2: What type of output do the evaluated system(s) generate? Select all that apply. If none match, select `Other' and describe.}}\label{sec:output}

\noindent\textit{Check-box options (select all that apply)}: 

\begin{enumerate}[itemsep=0cm,leftmargin=0.5cm,label={\small $\square$}]
            \item \egcvalue{raw/structured data}: numerical, symbolic, and other data, possibly structured into trees, graphs, graphical models, etc. May be the input e.g.\ to Referring Expression Generation (REG), end-to-end text generation, etc. {NB}: excludes linguistic structures.
            
            \item \egcvalue{deep linguistic representation (DLR)}: any of a variety of deep, underspecified, semantic representations, such as abstract meaning representations \citep[AMRs;][]{banarescu-etal-2013-abstract} or discourse representation structures \citep[DRSs;][]{kamp2013discourse}.
            
            \item \egcvalue{shallow linguistic representation (SLR)}: any of a variety of shallow, syntactic representations, e.g.\ Universal Dependency (UD) structures; typically the input to surface realisation.
            
            \item \egcvalue{text: subsentential unit of text}: a unit of text shorter than a sentence, e.g.\ Referring Expressions (REs), verb phrase, text fragment of any length; includes titles/headlines.
            
            \item \egcvalue{text: sentence}: a single sentence (or set of sentences).
            
            \item[\checkmark] \egcvalue{text: multiple sentences}: a sequence of multiple sentences, without any document structure (or a set of such sequences). 
            
            \item \egcvalue{text: document}: a text with document structure, such as a title, paragraph breaks or sections, e.g.\ a set of news reports for summarisation.
            
            \item \egcvalue{text: dialogue}: a dialogue of any length, excluding a single turn which would come under one of the other text types.
            
           \item \egcvalue{text: other}: select if output is text but doesn't match any of the above \textit{text:*} categories.
            
            \item \egcvalue{speech}: a recording of speech.
            
            \item \egcvalue{visual}: an image or video.
            
            \item \egcvalue{multi-modal}: catch-all value for any combination of data and/or linguistic representation and/or visual data etc.
            
            \item \egcvalue{human-generated `outputs'}: manually created stand-ins exemplifying outputs.
            
            \item \egcvalue{other (please specify)}: if output is none of the above, choose this option and describe it. 
            
        \end{enumerate}

\subsubsection*{\qsecbox{Question 2.3: How would you describe the task that the evaluated system(s) perform in mapping the inputs in Q2.1 to the outputs in Q2.2? Occasionally, more than one of the options below may apply. If none match, select `Other' and describe.}}\label{sec:task}

\noindent\textit{Check-box options (select all that apply)}:  

\begin{enumerate}[itemsep=0cm,leftmargin=0.5cm,label={\small $\square$}]
    \item \egcvalue{content selection/determination}: selecting  the specific content that will be expressed in the generated text from a representation of possible content. This could be attribute selection for REG (without the surface realisation step). Note that the output here is not text.
    
    \item \egcvalue{content ordering/structuring}: assigning an order and/or structure to content to be included in generated text. Note that the output here is not text.
    
    \item \egcvalue{aggregation}: converting inputs (typically \textit{deep linguistic representations} or \textit{shallow linguistic representations}) in some way in order to reduce redundancy (e.g.\  representations for `they like swimming', `they like running' $\rightarrow$ representation for `they like swimming and running').
   
    \item \egcvalue{referring expression generation}: generating \textit{text} to refer to a given referent, typically represented in the input as a set of attributes or a linguistic representation. 
    
    \item \egcvalue{lexicalisation}: associating (parts of) an input representation with specific lexical items to be used in their realisation. 
    
    \item[\checkmark] \egcvalue{deep generation}: one-step text generation from \textit{raw/structured data} or \textit{deep linguistic representations}. One-step means that no intermediate representations are passed from one independently run module to another.
    
    \item \egcvalue{surface realisation (SLR to text)}: one-step text generation from \textit{shallow linguistic representations}. One-step means that no intermediate representations are passed from one independently run module to another.
    
    \item \egcvalue{feature-controlled text generation}: generation of text that varies along specific dimensions where the variation is controlled via \textit{control feature}s specified as part of the input. Input is a non-textual representation (for feature-controlled text-to-text generation select the matching text-to-text task). 
    
    \item \egcvalue{data-to-text generation}: generation from \textit{raw/structured data} which may or may not include some amount of content selection as part of the generation process. Output is likely to be \textit{text:*} or \textit{multi-modal}.
    
    \item \egcvalue{dialogue turn generation}: generating a dialogue turn (can be a greeting or closing) from a representation of dialogue state and/or last turn(s), etc. 

    \item \egcvalue{question generation}: generation of questions from given input text and/or knowledge base such that the question can be answered from the input.
    
    \item \egcvalue{question answering}: input is a question plus optionally a set of reference texts and/or knowledge base, and the output is the answer to the question.
    
    \item[\checkmark] \egcvalue{paraphrasing/lossless simplification}: text-to-text generation where the aim is to preserve the meaning of the input while changing its wording. This can include the aim of changing the text on a given dimension, e.g.\ making it simpler, changing its stance or sentiment, etc., which may be controllable via input features. Note that this task type includes meaning-preserving text simplification (non-meaning preserving simplification comes under \textit{compression/lossy simplification} below).
    
    \item \egcvalue{compression/lossy simplification}: text-to-text generation that has the aim to generate a shorter, or shorter and simpler, version of the input text. This will normally affect meaning to some extent, but as a side effect, rather than the primary aim, as is the case in \textit{summarisation}.
    
    \item \egcvalue{machine translation}: translating text in a source language to text in a target language while maximally preserving the meaning. 
    
    \item \egcvalue{summarisation (text-to-text)}: output is an extractive or abstractive summary of the important/relevant/salient content of the input  document(s).

    \item \egcvalue{end-to-end text generation}: use this option if the single system task corresponds to more than one of tasks above, implemented either as separate modules pipelined together, or as one-step generation, other than \textit{deep generation} and \textit{surface realisation}.
    
    \item \egcvalue{image/video description}: input includes \textit{visual}, and the output describes it in some way.
    
    \item \egcvalue{post-editing/correction}: system edits and/or corrects the input text (typically itself the textual output from another system) to yield an improved version of the text.
       
    \item \egcvalue{other (please specify)}: if task is none of the above, choose this option and describe it.
    \end{enumerate}

\subsubsection*{\qsecbox{Question 2.4: Input Language(s), or `N/A'.}}
French.
        
\subsubsection*{\qsecbox{Question 2.5: Output Language(s), or `N/A'.}}
French.

\subsection{Output Sample, Evaluators, Experimental Design}\label{sec:design}

\subsubsection{Sample of system outputs (or human-authored stand-ins) evaluated (Questions 3.1.1--3.1.3)}

\subsubsection*{\qsecbox{Question 3.1.1: How many system outputs (or other evaluation items) are evaluated per system in the evaluation experiment? Answer should be an integer.}}
297.

\vspace{-.3cm}
\subsubsection*{\qsecbox{Question 3.1.2: How are system outputs (or other evaluation items) selected for inclusion in the evaluation experiment? If none match, select `Other' and describe.}}
\noindent\textit{Multiple-choice options (select one)}:  
\begin{enumerate}[itemsep=0cm,leftmargin=0.5cm,label={\LARGE $\circ$}]
    \item \egcvalue{by an automatic random process from a larger set}: outputs were selected for inclusion in the experiment by a script using a pseudo-random number generator; don't use this option if the script selects every $n$th output (which is not random). 
    \item \egcvalue{by an automatic random process but using stratified sampling over given properties}: use this option if selection was by a random script as above, but with added constraints ensuring that the sample is representative of the set of outputs it was selected from, in terms of given properties, such as sentence length, positive/negative stance, etc.
    \item \egcvalue{by manual, arbitrary selection}: output sample was selected by hand, or automatically from a manually compiled list, without a specific selection criterion.
    \item[\checkmark] \egcvalue{by manual selection aimed at achieving balance or variety relative to given properties}: selection by hand as above, but with specific selection criteria, e.g.\ same number of outputs from each time period.
    \item \egcvalue{Other (please specify)}: if selection method is none of the above, choose this option and describe it.
\end{enumerate}

\subsubsection*{\qsecbox{Question 3.1.3: What is the statistical power of the sample size?}}

\noindent Following the methodology of \citet{card-etal-2020-little}, we obtained a statistical power of 0.33 on the output sample w.r.t the automatic evaluation metrics, the two best-performing models (\texttt{JUDGEBERT-DA} and BERTScore). We used their online script to estimate the statistical power.

\subsubsection{Evaluators (Questions 3.2.1--3.2.4)}

\subsubsection*{\qsecbox{Question 3.2.1:  How many evaluators are there in this experiment? Answer should be an integer.}}

Five.

\subsubsection*{\qsecbox{Question 3.2.2:  What kind of evaluators are in this experiment? Select all that apply. If none match, select `Other' and describe. In all cases, provide details in the text box under `Other'.}}

\noindent\textit{Check-box options (select all that apply)}:  

\begin{enumerate}[itemsep=0cm,leftmargin=0.5cm,label={\small $\square$}]
    \item[\checkmark] \egcvalue{experts}: participants are considered domain experts, e.g.\ meteorologists evaluating a weather forecast generator, or nurses evaluating an ICU report generator.
    \item \egcvalue{non-experts}: participants are not domain experts.
    \item[\checkmark] \egcvalue{paid (including non-monetary compensation such as course credits)}: participants were given some form of compensation for their participation, including vouchers, course credits, and reimbursement for travel unless based on receipts.
    \item \egcvalue{not paid}: participants were not given compensation of any kind.
    \item \egcvalue{previously known to authors}: (one of the) researchers running the experiment knew some or all of the participants before recruiting them for the experiment.
    \item[\checkmark] \egcvalue{not previously known to authors}: none of the researchers running the experiment knew any of the participants before recruiting them for the experiment.
    \item \egcvalue{evaluators include one or more of the authors}: one or more researchers running the experiment was among the participants.
    \item[\checkmark]  \egcvalue{evaluators do not include any of the authors}: none of the researchers running the experiment were among the participants.
    \item \egcvalue{Other} (fewer than 4 of the above apply): we believe you should be able to tick 4 options of the above. If that's not the case, use this box to explain.
\end{enumerate}

\subsubsection*{\qsecbox{Question 3.2.3:  How are evaluators recruited?}}

Evaluators were recruited through a job offer on the University job board and interviewed prior to conducting the experiment.

\subsubsection*{\qsecbox{Question 3.2.4: What training and/or practice are evaluators given before starting on the evaluation itself?}}

First, the evaluators have been introduced to the task of text simplification
generation. 
They were then introduced to the dataset under study. 
They learned from an annotation guideline and practices on 15 examples before conducting the whole experiment. 
Evaluators did not need legal training since they all had domain background knowledge.

\subsection*{\qsecbox{Question 3.2.5:  What other characteristics do the evaluators have, known either because these were qualifying criteria, or from information gathered as part of the evaluation?}}

Evaluators have been selected based on their educational level, i.e. at least in their second year in law school, and interest in insurance law.

\subsubsection{Experimental design (Questions 3.3.1--3.3.8)}
\subsubsection*{\qsecbox{Question 3.3.1:  Has the experimental design been preregistered? If yes, on which registry?}}
No.

\subsubsection*{\qsecbox{Question 3.3.2: How are responses collected? E.g.\ paper forms, online survey tool, etc.}}

The answers were collected using a customized version of Prodigy\footnote{\url{https://prodi.gy/}}, hosted on Amazon Web Services.

\subsubsection*{\qsecbox{Question 3.3.3:  What quality assurance methods are used? Select all that apply. If none match, select `Other' and describe. In all cases, provide details in the text box under `Other'.}}

\noindent\textit{Check-box options (select all that apply)}:  
\vspace{-.1cm}

\begin{enumerate}[itemsep=0cm,leftmargin=0.5cm,label={\small $\square$}]
    \item[\checkmark] \egcvalue{evaluators are required to be native speakers of the language they evaluate}: mechanisms are in place to ensure all participants are native speakers of the language they evaluate.
    \item \egcvalue{automatic quality checking methods are used during/post evaluation}: evaluations are checked for quality by automatic scripts during or after evaluations, e.g.\ evaluators are given known bad/good outputs to check they're given bad/good scores on MTurk.
    \item[\checkmark] \egcvalue{manual quality checking methods are used during/post evaluation}: evaluations are checked for quality by a manual process  during or after evaluations, e.g.\ scores assigned by evaluators are monitored by researchers conducting the experiment.
    \item \egcvalue{evaluators are excluded if they fail quality checks (often or badly enough)}: there are conditions under which evaluations produced by participants are not included in the final results due to quality issues.
    \item \egcvalue{some evaluations are excluded because of failed quality checks}: there are conditions under which some (but not all) of the evaluations produced by some participants are not included in the final results due to quality issues.
    \item \egcvalue{none of the above}: tick this box if none of the above apply.
    \item \egcvalue{Other (please specify)}: use this box to describe any other quality assurance methods used during or after evaluations, and to provide additional details for any of the options selected above.
\end{enumerate}

\subsubsection*{\qsecbox{Question 3.3.4:  What do evaluators see when carrying out evaluations? Link to screenshot(s) and/or describe the evaluation interface(s).}}

When evaluating, evaluators see the input data (e.g., a complex sentence) and the simplification generated by the model. To reduce any bias toward public LLM (e.g., GPT4), they do not know the model name. They then independently provide a score for each generation.

\subsubsection*{\qsecbox{3.3.5: How free are evaluators regarding when and how quickly to carry out evaluations? Select all that apply. In all cases, provide details in the text box under `Other'.}}

\noindent\textit{Check-box options (select all that apply)}:  
\vspace{-.1cm}

\begin{enumerate}[itemsep=0cm,leftmargin=0.5cm,label={\small $\square$}]
    \item \egcvalue{evaluators have to complete each individual assessment within a set time}: evaluators are timed while carrying out each assessment and cannot complete the assessment once time has run out.
    \item \egcvalue{evaluators have to complete the whole evaluation in one sitting}: partial progress cannot be saved and the evaluation returned to on a later occasion.
    \item[\checkmark] \egcvalue{neither of the above}: Choose this option if neither of the above are the case in the experiment.
    \item \egcvalue{Other (please specify)}: Use this space to describe any other way in which time taken or number of sessions used by evaluators is controlled in the experiment, and to provide additional details for any of the options selected above.
\end{enumerate}

\subsubsection*{\qsecbox{3.3.6: Are evaluators told they can ask questions about the evaluation and/or provide feedback? Select all that apply. In all cases, provide details in the text box under `Other'.}}

\noindent\textit{Check-box options (select all that apply)}:  

\begin{enumerate}[itemsep=0cm,leftmargin=0.5cm,label={\small $\square$}]
    \item[\checkmark] \egcvalue{evaluators are told they can ask any questions during/after receiving initial training/instructions, and before the start of the evaluation}: evaluators are told explicitly that they can ask  questions about the evaluation experiment \textit{before} starting on their assessments, either during or after training.
    \item \egcvalue{evaluators are told they can ask any questions during the evaluation}: evaluators are told explicitly that they can ask  questions about the evaluation experiment \textit{during} their assessments.
    \item \egcvalue{evaluators are asked for feedback and/or comments after the evaluation, e.g.\ via an exit questionnaire or a comment box}: evaluators are explicitly asked to provide feedback and/or comments about the experiment \textit{after} their assessments, either verbally or in written form.
    \item \egcvalue{None of the above}: Choose this option if none of the above are the case in the experiment.
    \item \egcvalue{Other (please specify)}: use this space to describe any other ways you provide for evaluators to ask questions or provide feedback.
\end{enumerate}

\subsubsection*{\qsecbox{3.3.7: What are the experimental conditions in which evaluators carry out the evaluations? If none match, select `Other’ and describe.}}

\noindent\textit{Multiple-choice options (select one)}:  
\begin{enumerate}[itemsep=0cm,leftmargin=0.5cm,label={\LARGE $\circ$}]
    \item[\checkmark] \egcvalue{evaluation carried out by evaluators at a place of their own choosing, e.g.\ online, using a paper form, etc.}: evaluators are given access to the tool or form specified in Question 3.3.2, and subsequently choose where to carry out their evaluations.
    \item \egcvalue{evaluation carried out in a lab, and conditions are the same for each evaluator}: evaluations are carried out in a lab, and conditions in which evaluations are carried out \textit{are} controlled to be the same, i.e.\ the different evaluators all carry out the evaluations in identical conditions of quietness, same type of computer, same room, etc. Note we're not after very fine-grained differences here, such as time of day  or temperature, but the line is difficult to draw, so some judgment is involved here.
    \item \egcvalue{evaluation carried out in a lab, and conditions vary for different evaluators}: choose this option if evaluations are carried out in a lab, but the preceding option does not apply, i.e.\ conditions in which evaluations are carried out are \textit{not} controlled to be the same.
    \item \egcvalue{evaluation carried out in a real-life situation, and conditions are the same for each evaluator}: evaluations are carried out in a real-life situation, i.e.\ one that would occur whether or not the evaluation was carried out (e.g.\ evaluating a dialogue system deployed in a live chat function on a website), and conditions in which evaluations are carried out \textit{are} controlled to be the same. 
    \item \egcvalue{evaluation carried out in a real-life situation, and conditions vary for different evaluators}: choose this option if evaluations are carried out in a real-life situation, but the preceding option does not apply, i.e.\ conditions in which evaluations are carried out are \textit{not} controlled to be the same.
    \item \egcvalue{evaluation carried out outside of the lab, in a situation designed to resemble a real-life situation, and conditions are the same for each evaluator}: evaluations are carried out outside of the lab, in a situation intentionally similar to a real-life situation (but not actually a real-life situation), e.g.\ user-testing a navigation system where the destination is part of the evaluation design, rather than chosen by the user. Conditions in which evaluations are carried out \textit{are} controlled to be the same. 
    \item \egcvalue{evaluation carried out outside of the lab, in a situation designed to resemble a real-life situation, and conditions vary for different evaluators}: choose this option if evaluations are carried out outside of the lab, in a situation intentionally similar to a real-life situation, but the preceding option does not apply, i.e.\ conditions in which evaluations are carried out are \textit{not} controlled to be the same. 
    \item \egcvalue{Other (please specify)}: Use this space to provide additional, or alternative, information about the conditions in which evaluators carry out assessments, not covered by the options above.
\end{enumerate}

\subsubsection*{\qsecbox{3.3.8:  Unless the evaluation is carried out at a place of the evaluators'  own choosing, briefly describe the (range of different) conditions in which evaluators carry out the evaluations.}}
N/A.

\subsection{Quality Criterion \textit{n} -- Definition and Operationalisation}
\label{sec:criteria}

\subsubsection{Quality criterion properties (Questions 4.1.1--4.1.3)}
\subsubsection*{\qsecbox{Question 4.1.1:  What type of quality is assessed by the quality criterion?}}

\noindent\textit{Multiple-choice options (select one)}:  
\begin{enumerate}[itemsep=0cm,leftmargin=0.5cm,label={\LARGE $\circ$}]
    \item[\checkmark] \egcvalue{Correctness}: select this option if it is possible to state,  generally for all outputs,  the conditions under which outputs are maximally correct (hence of maximal quality).  E.g.\ for Grammaticality, outputs are (maximally) correct if they contain no grammatical errors; for Semantic Completeness, outputs are correct if they express all the content in the input.
    \item \egcvalue{Goodness}: select this option if, in contrast to correctness criteria, there is no single, general mechanism for deciding when outputs are maximally good, only for deciding for two outputs which is better and which is worse. E.g.\ for Fluency, even if outputs contain no disfluencies, there may be other ways in which any given output could be more fluent.
    \item \egcvalue{Features}: choose this option if, in terms of property $X$ captured by the criterion, outputs are not generally better if they are more $X$, but instead, depending on evaluation context, more $X$ may be better or less $X$ may be better. E.g.\ outputs can be more specific or less specific, but it’s not the case that outputs are, in the general case, better when they are more specific.
\end{enumerate}

\subsubsection*{\qsecbox{Question 4.1.2:  Which aspect of system outputs is assessed by the quality criterion?}}

\noindent\textit{Multiple-choice options (select one)}:  
\begin{enumerate}[itemsep=0cm,leftmargin=0.5cm,label={\LARGE $\circ$}]
    \item \egcvalue{Form of output}: choose this option if the criterion assesses the form of outputs alone, e.g.\ Grammaticality is only about the form, a sentence can be grammatical yet be wrong or nonsensical in terms of content.
    \item[\checkmark] \egcvalue{Content of output}: choose this option if the criterion assesses the content/meaning of the output alone, e.g.\ Meaning Preservation only assesses output content; two sentences can be considered to have the same meaning, but differ in form.
    \item \egcvalue{Both form and content of output}: choose this option if the criterion assesses outputs as a whole, not just form or just content. E.g.\ Coherence is a property of outputs as a whole, either form or meaning can detract from it.
\end{enumerate}

\subsubsection*{\qsecbox{Question 4.1.3: Is each output assessed for quality in its own right, or with reference to a system-internal or external frame of reference?}}

\noindent\textit{Multiple-choice options (select one)}:  

\begin{enumerate}[itemsep=0cm,leftmargin=0.5cm,label={\LARGE $\circ$}]
    \item \egcvalue{Quality of output in its own right}: choose this option if output quality is assessed without referring to anything other than the output itself, i.e.\ no  system-internal or external frame of reference. E.g.\ Poeticness is assessed by considering (just) the output and how poetic it is. 
    \item[\checkmark] \egcvalue{Quality of output relative to the input}:  choose this option if output quality is assessed  relative to the input. E.g.\ Answerability is the degree to which the output question can be answered from information in the input.
    \item \egcvalue{Quality of output relative to a system-external frame of reference}: choose this option if output quality is assessed with reference to system-external information, such as a knowledge base, a person’s individual writing style, or the performance of an embedding system. E.g.\ Factual Accuracy assesses outputs relative to a source of real-world knowledge.
\end{enumerate}

\subsubsection{Evaluation mode properties (Questions 4.2.1--4.2.3)}

Questions 4.2.1--4.2.3 record properties that are orthogonal to quality criteria, i.e.\ any given quality criterion can in principle be combined with any of the modes (although some combinations are more common than others). 

\subsubsection*{\qsecbox{Question 4.2.1: Does an individual assessment involve an objective or a subjective judgment?}}

\noindent\textit{Multiple-choice options (select one)}:  

\begin{enumerate}[itemsep=0cm,leftmargin=0.5cm,label={\LARGE $\circ$}]
    \item[\checkmark] \egcvalue{Objective}: Examples of objective assessment include any automatically counted or otherwise quantified measurements such as mouse-clicks, occurrences in text, etc. Repeated assessments of the same output with an objective-mode evaluation method always yield the same score/result.
    \item \egcvalue{Subjective}: Subjective assessments involve ratings, opinions and preferences by evaluators. Some criteria lend themselves more readily to subjective assessments, e.g.\ Friendliness of a conversational agent, but an objective measure e.g.\ based on lexical markers is also conceivable.
\end{enumerate}
    
\subsubsection*{\qsecbox{Question 4.2.2: Are outputs assessed in absolute or relative terms?}}

\noindent\textit{Multiple-choice options (select one)}:  
\begin{enumerate}[itemsep=0cm,leftmargin=0.5cm,label={\LARGE $\circ$}]
    \item[\checkmark] \egcvalue{Absolute}: choose this option if evaluators are shown outputs from a single system during each individual assessment. 
    \item \egcvalue{Relative}: choose this option if evaluators are shown outputs from multiple systems at the same time during assessments, typically ranking or preference-judging them.
\end{enumerate}

\subsubsection*{\qsecbox{Question 4.2.3: Is the evaluation intrinsic or extrinsic?}}

\noindent\textit{Multiple-choice options (select one)}:  
\begin{enumerate}[itemsep=0cm,leftmargin=0.5cm,label={\LARGE $\circ$}]
    \item \egcvalue{Intrinsic}: Choose this option if quality of outputs is assessed \textit{without} considering their \textit{effect} on something external to the system, e.g.\ the performance of an embedding system or of a user at a task.
    \item[\checkmark] \egcvalue{Extrinsic}: Choose this option if quality of outputs is assessed in terms of their \textit{effect} on something external to the system such as the performance of an embedding system or of a user at a task.
\end{enumerate}

\subsubsection{Response elicitation (Questions 4.3.1--4.3.11)}

\subsubsection*{\qsecbox{Question 4.3.1: What do you call the quality criterion in explanations/interfaces to evaluators?  Enter `N/A' if criterion not named.}}
Legal meaning.

\subsubsection*{\qsecbox{Question 4.3.2:  What definition do you give for the quality criterion in explanations/interfaces to evaluators? Enter `N/A' if no definition given.}}
We define \textbf{legal meaning} as \guillemet{[the] measures [of] how well the output text conveys the legal details and exceptions and does not misrepresent the law}.
 
\subsubsection*{\qsecbox{Question 4.3.3:  Size of scale or other rating instrument (i.e.\ how many different possible values there are). Answer should be an integer or `continuous' (if it's not possible to state how many possible responses there are). Enter `N/A' if there is no rating instrument.}}
10.

\subsubsection*{\qsecbox{Question 4.3.4: List or range of possible values of the scale or other rating instrument. Enter `N/A', if there is no rating instrument.}} 
1, 2, 3, 4, 5, 6, 7, 8, 9, 10.

\subsubsection*{\qsecbox{Question 4.3.5:  How is the scale or other rating instrument presented to evaluators? If none match, select `Other’ and describe.}}

\noindent\textit{Multiple-choice options (select one)}:  

\begin{enumerate}[itemsep=0cm,leftmargin=0.5cm,label={\LARGE $\circ$}]
    \item \egcvalue{Multiple-choice options}: choose this option if evaluators select exactly one of multiple options.
    \item \egcvalue{Check-boxes}: choose this option if evaluators select any number of options from multiple given options.
    \item \egcvalue{Slider}: choose this option if evaluators move a pointer on a slider scale to the position corresponding to their assessment.
    \item \egcvalue{N/A (there is no rating instrument)}: choose this option if there is no rating instrument.
    \item[\checkmark] \egcvalue{Other (please specify)}: choose this option if there is a rating instrument, but none of the above adequately describe the way you present it to evaluators. Use the text box to describe the rating instrument and link to a screenshot.
\end{enumerate}

Due to Prodigy's limitations regarding their slider component (only one per page), we used a free-form text box. Since we have few highly skilled evaluators, collecting data was not a problem.

\subsubsection*{\qsecbox{Question 4.3.6:  If there is no rating instrument, describe briefly what task the evaluators perform (e.g.\ ranking multiple outputs, finding information, playing a game, etc.), and what information is recorded. Enter `N/A' if there is a rating instrument.}}
N/A.

\subsubsection*{\qsecbox{Question 4.3.7:  What is the verbatim question, prompt or instruction given to evaluators (visible to them during each individual assessment)?}}

Here is the verbatim question and instruction in French to evaluators (in the following list), we also present an automatic translation of these instruction in the second list.

\begin{enumerate}[leftmargin=*, label={}, noitemsep]
    \item \textit{\textbf{Étape 1 de l'évaluation:} Comment évaluez-vous le niveau de difficulté du texte généré par le modèle?}
    \item \textit{\textbf{Étape 2 de l'évaluation:} Selon vous, quelle est la qualification du texte?}
    \item \textit{\textbf{Étape 3 de l'évaluation: }Selon vous, quel est le niveau de précision légale du texte généré par le modèle sur une échelle de 1 à 10?}
    \item \textit{\textbf{Commentaires (si applicable):}}
\end{enumerate}

Here is the automatic English translation of the verbatim question and instructions for evaluators.
\begin{enumerate}[leftmargin=*, label={}, noitemsep]
    \item \textbf{Evaluation step 1:} How would you rate the level of difficulty of the text generated by the template?
    \item \textbf{Evaluation step 2:} How do you rate the text?
    \item \textbf{Evaluation step 3:} What do you think is the legal accuracy level of the text generated by the model on a scale of 1 to 10?
    \item \textbf{Comments (if applicable):}
\end{enumerate}

\subsubsection*{\qsecbox{Question 4.3.8:  Form of response elicitation. If none match, select `Other' and describe.}}
\noindent\textit{Multiple-choice options (select one)}:\footnote{Explanations adapted from \citet{howcroft2019typology}.}

\begin{enumerate}[itemsep=0cm,leftmargin=0.5cm,label={\LARGE $\circ$}]
    \item \egcvalue{(dis)agreement with quality statement}: Participants specify the degree to which they agree with a given quality statement by indicating their agreement on a rating instrument. The rating instrument is labelled with degrees of agreement and can additionally have numerical labels.  E.g.\ \textit{This text is fluent --- 1=strongly disagree...5=strongly agree}.
    \item \egcvalue{direct quality estimation}: Participants are asked to provide a rating using a rating instrument, which typically (but not always) mentions the quality criterion explicitly. E.g.\ \textit{How fluent is this text? --- 1=not at all fluent...5=very fluent}.
    \item \egcvalue{relative quality estimation (including ranking)}: Participants evaluate two or more items in terms of which is better.
    E.g.\ \textit{Rank these texts in terms of fluency}; \textit{Which of these texts is more fluent?}; \textit{Which of these items do you prefer?}.
    \item[\checkmark] \egcvalue{counting occurrences in text}: Evaluators are asked to count how many times some type of phenomenon occurs, e.g.\ the number of facts contained in the output that are inconsistent with the input.
    \item \egcvalue{qualitative feedback (e.g.\ via comments entered in a text box)}: Typically, these are responses to open-ended questions in a survey or interview.
    \item \egcvalue{evaluation through post-editing/annotation}: Choose this option if the evaluators' task consists of editing or inserting annotations in text. E.g.\ evaluators may perform error correction and edits are then automatically measured to yield a numerical score.
    \item \egcvalue{output classification or labelling}: Choose this option if evaluators assign outputs to categories. E.g.\ \textit{What is the overall sentiment of this piece of text? --- Positive/neutral/negative.}
    \item \egcvalue{user-text interaction measurements}: choose this option if participants in the evaluation experiment interact with a text in some way, and measurements are taken of their interaction. E.g.\ reading speed, eye movement tracking, comprehension questions, etc. Excludes situations where participants are given a task to solve and their performance is measured which comes under the next option.
    \item \egcvalue{task performance measurements}: choose this option if participants in the evaluation experiment are given a task to perform, and measurements are taken of their performance at the task.  E.g.\ task is finding information, and task performance measurement is task completion speed and success rate.
    \item \egcvalue{user-system interaction measurements}: choose this option if participants in the evaluation experiment interact with a system in some way, while measurements are taken of their interaction. E.g.\ duration of interaction, hyperlinks followed, number of likes, or completed sales.
    \item \egcvalue{Other (please specify)}: Use the text box to describe the form of response elicitation used in assessing the quality criterion if it doesn't fall in any of the above categories.
\end{enumerate}

\subsubsection*{\qsecbox{Question 4.3.9:  How are raw responses from participants aggregated or otherwise processed to obtain reported scores for this quality criterion? State if no scores reported.}}

Macro averages are computed from numerical scores to provide a summary.

\subsubsection*{\qsecbox{Question 4.3.10:  Method(s) used for determining effect size and significance of findings for this quality criterion.}}

\noindent\textit{What to enter in the text box}: A list of  methods used for calculating the effect size and significance of any results, both as reported in the paper given in Question 1.1, for this quality criterion. If none calculated, state `None'.

\bigskip
\noindent None.

\vspace{-.3cm}
\subsubsection*{\qsecbox{Question 4.3.11:  Has the inter-annotator and intra-annotator agreement between evaluators for this quality criterion been measured? If yes, what method was used, and what are the agreement scores?}}

Krippendorff's alpha \cite{hayes2007answering} is used to measure inter-annotator agreement. Krippendorff's alpha are detailled in \autoref{tab:annotators_res}.

\subsection{Ethics}\label{sec:ethics}

\subsubsection*{\qsecbox{Question 5.1: Has the evaluation experiment this sheet is being completed for, or the larger study it is part of, been approved by a research ethics committee? If yes, which research ethics committee?}}
No.
\vspace{-0.4cm}

\subsubsection*{\qsecbox{Question 5.2: Do any of the system outputs (or human-authored stand-ins) evaluated, or do any of the responses collected, in the experiment contain personal data (as defined in GDPR Art. 4, §1: https://gdpr.eu/article-4-definitions/)? If yes, describe data and state how addressed.}}
No.
\vspace{-0.4cm}

\subsubsection*{\qsecbox{Question 5.3: Do any of the system outputs (or human-authored stand-ins) evaluated, or do any of the responses collected, in the experiment contain special category information (as defined in GDPR Art. 9, §1: https://gdpr.eu/article-9-processing-special-categories-of-personal-data-prohibited/)? If yes, describe data and state how addressed.}}
No.
\vspace{-0.4cm}

\subsubsection*{\qsecbox{Question 5.4:  Have any impact assessments been carried out for the evaluation experiment, and/or any data collected/evaluated in connection with it? If yes, summarise approach(es) and outcomes.}}
No.
\vspace{-0.4cm}

%% file: article.bbl
\begin{thebibliography}{69}
\expandafter\ifx\csname natexlab\endcsname\relax\def\natexlab#1{#1}\fi

\bibitem[{Agrawal and Carpuat(2024)}]{agrawal2024text}
Sweta Agrawal and Marine Carpuat. 2024.
\newblock {Do Text Simplification Systems Preserve Meaning? A Human Evaluation
  via Reading Comprehension}.
\newblock \emph{Transactions of the Association for Computational Linguistics},
  12:432--448.

\bibitem[{Alva-Manchego et~al.(2020)Alva-Manchego, Martin, Bordes, Scarton,
  Sagot, and Specia}]{alva2020asset}
Fernando Alva-Manchego, Louis Martin, Antoine Bordes, Carolina Scarton,
  Beno{\^\i}t Sagot, and Lucia Specia. 2020.
\newblock {ASSET: A Dataset for Tuning and Evaluation of Sentence
  Simplification Models With Multiple Rewriting Transformations}.
\newblock In \emph{Annual Meeting of the Association for Computational
  Linguistics}, pages 4668–--4679.

\bibitem[{AMF(2014)}]{amf}
Autorité des marchés~financiers AMF. 2014.
\newblock \href
  {https://lautorite.qc.ca/fileadmin/lautorite/formulaires/professionnels/assureurs/automobile/fpq_1.pdf}
  {Formulaire de police d’assurance automobile du québec}.
\newblock Accessed: 2024-05-12.

\bibitem[{Antoun et~al.(2024)Antoun, Kulumba, Touchent, Éric de~la Clergerie,
  Sagot, and Seddah}]{antoun2024camembert20smarterfrench}
Wissam Antoun, Francis Kulumba, Rian Touchent, Éric de~la Clergerie, Benoît
  Sagot, and Djamé Seddah. 2024.
\newblock \href {http://arxiv.org/abs/2411.08868} {{CamemBERT 2.0: A Smarter
  French Language Model Aged to Perfection}}.

\bibitem[{Aumiller et~al.(2022)Aumiller, Chouhan, and Gertz}]{aumiller2022eur}
Dennis Aumiller, Ashish Chouhan, and Michael Gertz. 2022.
\newblock {EUR-Lex-Sum: A Multi-and Cross-lingual Dataset for Long-form
  Summarization in the Legal Domain}.
\newblock \emph{arXiv:2210.13448}.

\bibitem[{Banarescu et~al.(2013)Banarescu, Bonial, Cai, Georgescu, Griffitt,
  Hermjakob, Knight, Koehn, Palmer, and
  Schneider}]{banarescu-etal-2013-abstract}
Laura Banarescu, Claire Bonial, Shu Cai, Madalina Georgescu, Kira Griffitt, Ulf
  Hermjakob, Kevin Knight, Philipp Koehn, Martha Palmer, and Nathan Schneider.
  2013.
\newblock \href {https://aclanthology.org/W13-2322} {{A}bstract {M}eaning
  {R}epresentation for sembanking}.
\newblock In \emph{Proceedings of the 7th Linguistic Annotation Workshop and
  Interoperability with Discourse}, pages 178--186, Sofia, Bulgaria.
  Association for Computational Linguistics.

\bibitem[{Batra et~al.(2021)Batra, Jain, Heidari, Arun, Youngs, Li, Donmez,
  Mei, Kuo, Bhardwaj, Kumar, and White}]{batra-etal-2021-building}
Soumya Batra, Shashank Jain, Peyman Heidari, Ankit Arun, Catharine Youngs,
  Xintong Li, Pinar Donmez, Shawn Mei, Shiunzu Kuo, Vikas Bhardwaj, Anuj Kumar,
  and Michael White. 2021.
\newblock \href {https://doi.org/10.18653/v1/2021.emnlp-main.53} {Building
  adaptive acceptability classifiers for neural {NLG}}.
\newblock In \emph{Proceedings of the 2021 Conference on Empirical Methods in
  Natural Language Processing}, pages 682--697, Online and Punta Cana,
  Dominican Republic. Association for Computational Linguistics.

\bibitem[{Beauchemin and Khoury(2023)}]{Beauchemin2023RISC}
David Beauchemin and Richard Khoury. 2023.
\newblock {RISC}: Generating {Realistic} {Synthetic} {Bilingual} {Insurance}
  {Contract}.
\newblock \emph{Proceedings of the Canadian Conference on Artificial
  Intelligence}.
\newblock Https://caiac.pubpub.org/pub/k18zu6c9.

\bibitem[{Beauchemin et~al.(2023)Beauchemin, Saggion, and
  Khoury}]{beauchemin2023meaningbert}
David Beauchemin, Horacio Saggion, and Richard Khoury. 2023.
\newblock {MeaningBERT: Assessing Meaning Preservation Between Sentences}.
\newblock \emph{Frontiers in Artificial Intelligence}, 6.

\bibitem[{Bender et~al.(2021)Bender, Gebru, McMillan-Major, and
  Shmitchell}]{bender2021dangers}
Emily~M Bender, Timnit Gebru, Angelina McMillan-Major, and Shmargaret
  Shmitchell. 2021.
\newblock {On the Dangers of Stochastic Parrots: Can Language Models Be Too
  Big?}
\newblock In \emph{Proceedings of the ACM conference on fairness,
  accountability, and transparency}, pages 610--623.

\bibitem[{BIC(2009)}]{bac}
Bureau d'assurance du~Canada BIC. 2009.
\newblock \href
  {https://archives.bape.gouv.qc.ca/sections/mandats/Gaz_de_schiste/documents/DQ1.1.1_%20F.pdf}
  {Formulaire d’assurance habitation du québec}.
\newblock Accessed: 2024-05-12.

\bibitem[{Card et~al.(2020)Card, Henderson, Khandelwal, Jia, Mahowald, and
  Jurafsky}]{card-etal-2020-little}
Dallas Card, Peter Henderson, Urvashi Khandelwal, Robin Jia, Kyle Mahowald, and
  Dan Jurafsky. 2020.
\newblock \href {https://doi.org/10.18653/v1/2020.emnlp-main.745} {With little
  power comes great responsibility}.
\newblock In \emph{Proceedings of the 2020 Conference on Empirical Methods in
  Natural Language Processing (EMNLP)}, pages 9263--9274, Online. Association
  for Computational Linguistics.

\bibitem[{Cardon and Grabar(2020)}]{cardon2020french}
R{\'e}mi Cardon and Natalia Grabar. 2020.
\newblock {French Biomedical Text Simplification: When Small and Precise
  Helps}.
\newblock In \emph{Proceedings of the International Conference on Computational
  Linguistics}, pages 710--716.

\bibitem[{Caron(2024)}]{assurance_book}
Vincent Caron. 2024.
\newblock \emph{{Assurance, biens et responsabilité}}.
\newblock Édition Yvon Blais, Montréal.
\newblock Préface de l’honorable Pierre J. Dalphond.

\bibitem[{Douka et~al.(2021)Douka, Abdine, Vazirgiannis, El~Hamdani, and
  Amariles}]{douka2021juribert}
Stella Douka, Hadi Abdine, Michalis Vazirgiannis, Rajaa El~Hamdani, and
  David~Restrepo Amariles. 2021.
\newblock {JuriBERT: A Masked-Language Model Adaptation for French Legal Text}.
\newblock In \emph{Natural Legal Language Processing Workshop}, pages 95--101.
  Association for Computational Linguistics.

\bibitem[{Feng et~al.(2023)Feng, Qiang, Li, Yuan, and Zhu}]{feng2023sentence}
Yutao Feng, Jipeng Qiang, Yun Li, Yunhao Yuan, and Yi~Zhu. 2023.
\newblock {Sentence Simplification via Large Language Models}.
\newblock \emph{arXiv:2302.11957}.

\bibitem[{Fr{\'e}chette(2010)}]{frechette2010qualification}
Pascal Fr{\'e}chette. 2010.
\newblock La qualification des contrats: aspects th{\'e}oriques.
\newblock \emph{Les Cahiers de droit}, 51(1):117--158.

\bibitem[{Gala et~al.(2020)Gala, Tack, Javourey-Drevet, Fran{\c{c}}ois, and
  Ziegler}]{gala2020alector}
N{\'u}ria Gala, Ana{\"\i}s Tack, Ludivine Javourey-Drevet, Thomas
  Fran{\c{c}}ois, and Johannes~C Ziegler. 2020.
\newblock Alector: A parallel corpus of simplified french texts with alignments
  of misreadings by poor and dyslexic readers.
\newblock In \emph{Proceedings of the Language Resources and Evaluation
  Conference}, pages 1353--1361.

\bibitem[{Garimella et~al.(2022)Garimella, Sancheti, Aggarwal, Ganesh, Chhaya,
  and Kambhatla}]{garimella2022text}
Aparna Garimella, Abhilasha Sancheti, Vinay Aggarwal, Ananya Ganesh, Niyati
  Chhaya, and Nanda Kambhatla. 2022.
\newblock {Text Simplification for Legal Domain: Insights and Challenges}.
\newblock In \emph{Proceedings of the Natural Legal Language Processing
  Workshop}, pages 296--304.

\bibitem[{Garneau et~al.(2021)Garneau, Gaumond, Lamontagne, and
  D\'{e}ziel}]{10.1145/3462757.3466147}
Nicolas Garneau, Eve Gaumond, Luc Lamontagne, and Pierre-Luc D\'{e}ziel. 2021.
\newblock \href {https://doi.org/10.1145/3462757.3466147} {{CriminelBART: A
  French Canadian Legal Language Model Specialized in Criminal Law}}.
\newblock In \emph{Proceedings of the International Conference on Artificial
  Intelligence and Law}, ICAIL '21, page 256–257, New York, NY, USA.
  Association for Computing Machinery.

\bibitem[{Garneau et~al.(2022)Garneau, Gaumond, Lamontagne, and
  D{\'e}ziel}]{garneau-etal-2022-evaluating}
Nicolas Garneau, Eve Gaumond, Luc Lamontagne, and Pierre-Luc D{\'e}ziel. 2022.
\newblock \href {https://doi.org/10.18653/v1/2022.inlg-main.7} {Evaluating
  legal accuracy of neural generators on the generation of criminal court
  dockets description}.
\newblock In \emph{Proceedings of the International Conference on Natural
  Language Generation}, pages 73--99, Waterville, Maine, USA and virtual
  meeting. Association for Computational Linguistics.

\bibitem[{Gatt and Krahmer(2018)}]{gatt2018survey}
Albert Gatt and Emiel Krahmer. 2018.
\newblock {Survey of the State of the Art in Natural Language Generation: Core
  Tasks, Applications and Evaluation}.
\newblock \emph{Journal of Artificial Intelligence Research}, 61:65--170.

\bibitem[{Grabar and Cardon(2018)}]{grabar2018clear}
Natalia Grabar and R{\'e}mi Cardon. 2018.
\newblock {Clear-Simple Corpus for Medical French}.
\newblock In \emph{ATA}.

\bibitem[{Hagan(2023)}]{hagan2023good}
Margaret Hagan. 2023.
\newblock {Good AI Legal Help, Bad AI Legal Help: Establishing Quality
  Standards for Responses to People’s Legal Problem Stories}.
\newblock In \emph{JURIX}, volume 2023.

\bibitem[{Hayes and Krippendorff(2007)}]{hayes2007answering}
Andrew~F Hayes and Klaus Krippendorff. 2007.
\newblock {Answering the Call for a Standard Reliability Measure for Coding
  Data}.
\newblock \emph{Communication methods and measures}, 1(1):77--89.

\bibitem[{Hendrycks et~al.(2021)Hendrycks, Burns, Chen, and
  Ball}]{hendrycks2021cuad}
Dan Hendrycks, Collin Burns, Anya Chen, and Spencer Ball. 2021.
\newblock {CUAD: An Expert-Annotated NLP Dataset for Legal Contract Review}.
\newblock In \emph{Conference on Neural Information Processing Systems Datasets
  and Benchmarks Track}.

\bibitem[{Honnibal et~al.(2020)Honnibal, Montani, Van~Landeghem, and
  Boyd}]{Honnibal_spaCy_Industrial-strength_Natural_2020}
Matthew Honnibal, Ines Montani, Sofie Van~Landeghem, and Adriane Boyd. 2020.
\newblock {SpaCy: Industrial-strength Natural Language Processing in Python}.

\bibitem[{Howcroft and Bergvall-K{\aa}reborn(2019)}]{howcroft2019typology}
Debra Howcroft and Birgitta Bergvall-K{\aa}reborn. 2019.
\newblock {A Typology of Crowdwork Platforms}.
\newblock \emph{Work, employment and society}, 33(1):21--38.

\bibitem[{Hu et~al.(2024)Hu, Chen, Du, Peng, Keloth, Zuo, Zhou, Li, Jiang, Lu
  et~al.}]{hu2024improving}
Yan Hu, Qingyu Chen, Jingcheng Du, Xueqing Peng, Vipina~Kuttichi Keloth,
  Xu~Zuo, Yujia Zhou, Zehan Li, Xiaoqian Jiang, Zhiyong Lu, et~al. 2024.
\newblock {Improving Large Language Models for Clinical Named Entity
  Recognition via Prompt Engineering}.
\newblock \emph{Journal of the American Medical Informatics Association}, page
  ocad259.

\bibitem[{James et~al.(2013)James, Witten, Hastie, and
  Tibshirani}]{james2013introduction}
Gareth James, Daniela Witten, Trevor Hastie, and Robert Tibshirani. 2013.
\newblock \emph{{An Introduction to Statistical Learning}}, volume 112.
\newblock Springer.

\bibitem[{Janatian et~al.(2023)Janatian, Westermann, Tan, Savelka, and
  Benyekhlef}]{sileno2023text}
Samyar Janatian, Hannes Westermann, Jinzhe Tan, Jaromir Savelka, and Karim
  Benyekhlef. 2023.
\newblock {From Text to Structure: Using Large Language Models to Support the
  Development of Legal Expert Systems}.
\newblock \emph{Legal Knowledge and Information Systems}, pages 167--176.

\bibitem[{Kamp and Reyle(2013)}]{kamp2013discourse}
Hans Kamp and Uwe Reyle. 2013.
\newblock \emph{{From Discourse to Logic: Introduction to Modeltheoretic
  Semantics of Natural Language, Formal Logic and Discourse Representation
  Theory}}, volume~42.
\newblock Springer Science \& Business Media.

\bibitem[{Kandel and Moles(1958)}]{kandel1958application}
Liliane Kandel and Abraham Moles. 1958.
\newblock Application de l’indice de flesch {\`a} la langue fran{\c{c}}aise.
\newblock \emph{Cahiers Etudes de Radio-T{\'e}l{\'e}vision}, 19(1958):253--274.

\bibitem[{Kapoor et~al.(2024)Kapoor, Henderson, and
  Narayanan}]{kapoor2024promises}
Sayash Kapoor, Peter Henderson, and Arvind Narayanan. 2024.
\newblock {Promises and Pitfalls of Artificial Intelligence for Legal
  Applications}.
\newblock \emph{Forthcoming in the Journal of Cross-disciplinary Research in
  Computational Law (CRCL)}.

\bibitem[{Katz et~al.(2023)Katz, Hartung, Gerlach, Jana, and
  Bommarito}]{katz2023natural}
Daniel~Martin Katz, Dirk Hartung, Lauritz Gerlach, Abhik Jana, and
  Michael~James Bommarito. 2023.
\newblock {Natural Language Processing in the Legal Domain}.
\newblock \emph{Available at SSRN 4336224}.

\bibitem[{Kew et~al.(2023)Kew, Chi, V{\'a}squez-Rodr{\'\i}guez, Agrawal,
  Aumiller, Alva-Manchego, and Shardlow}]{kew2023bless}
Tannon Kew, Alison Chi, Laura V{\'a}squez-Rodr{\'\i}guez, Sweta Agrawal, Dennis
  Aumiller, Fernando Alva-Manchego, and Matthew Shardlow. 2023.
\newblock {BLESS: Benchmarking Large Language Models on Sentence
  Simplification}.
\newblock \emph{arXiv:2310.15773}.

\bibitem[{Kumar et~al.(2023)Kumar, Balachandran, Njoo, Anastasopoulos, and
  Tsvetkov}]{kumar-etal-2023-mitigating}
Sachin Kumar, Vidhisha Balachandran, Lucille Njoo, Antonios Anastasopoulos, and
  Yulia Tsvetkov. 2023.
\newblock \href {https://doi.org/10.18653/v1/2023.emnlp-tutorial.5} {Mitigating
  societal harms in large language models}.
\newblock In \emph{Proceedings of the Conference on Empirical Methods in
  Natural Language Processing: Tutorial Abstracts}, pages 26--33, Singapore.
  Association for Computational Linguistics.

\bibitem[{Laban et~al.(2020)Laban, Hsi, Canny, and Hearst}]{laban2021summary}
Philippe Laban, Andrew Hsi, John Canny, and Marti~A Hearst. 2020.
\newblock {The Summary Loop: Learning to Write Abstractive Summaries Without
  Examples}.
\newblock In \emph{Proceedings of the Annual Meeting of the Association for
  Computational Linguistics}, pages 5135--5150.

\bibitem[{Lewis and White(2023)}]{lewis-white-2023-mitigating}
Ashley Lewis and Michael White. 2023.
\newblock \href {https://aclanthology.org/2023.tllm-1.4} {Mitigating harms of
  {LLM}s via knowledge distillation for a virtual museum tour guide}.
\newblock In \emph{Proceedings of the Workshop on Taming Large Language Models:
  Controllability in the era of Interactive Assistants!}, pages 31--45, Prague,
  Czech Republic. Association for Computational Linguistics.

\bibitem[{Likert(1932)}]{Likert1932ATF}
Rensis Likert. 1932.
\newblock \emph{{A Technique for the Measurement of Attitudes}}.
\newblock Archives of psychology ; no. 140. [s.n.], New York.

\bibitem[{Maddela et~al.(2022)Maddela, Dou, Heineman, and Xu}]{maddela2022lens}
Mounica Maddela, Yao Dou, David Heineman, and Wei Xu. 2022.
\newblock {LENS: A Learnable Evaluation Metric for Text Simplification}.
\newblock \emph{arXiv:2212.09739}.

\bibitem[{Madina et~al.(2024)Madina, Gonzalez-Dios, and
  Siegel}]{madina2024preliminary}
Margot Madina, Itziar Gonzalez-Dios, and Melanie Siegel. 2024.
\newblock {A Preliminary Study of ChatGPT for Spanish E2R Text Adaptation}.
\newblock In \emph{Proceedings of the Joint International Conference on
  Computational Linguistics, Language Resources and Evaluation}, pages
  1422--1434.

\bibitem[{{Moffatt \textit{v. Air Canada}}(2024)}]{air_Canada_court}
{Moffatt \textit{v. Air Canada}}. 2024.
\newblock Civil Resolution Tribunal of British Columbia 149.

\bibitem[{Montani and Honnibal(2018)}]{prodigy_montani_honnibal}
Ines Montani and Matthew Honnibal. 2018.
\newblock \href {https://prodi.gy/} {{Prodigy: A Modern and Scriptable
  Annotation Tool for Creating Training Data for Machine Learning Models}}.

\bibitem[{Mosbach et~al.(2021)Mosbach, Andriushchenko, and
  Klakow}]{mosbach2020stability}
Marius Mosbach, Maksym Andriushchenko, and Dietrich Klakow. 2021.
\newblock {On the Stability of Fine-Tuning BERT: Misconceptions, Explanations,
  and Strong Baselines}.
\newblock In \emph{International Conference on Learning Representations}.

\bibitem[{Nozza et~al.(2023)Nozza, Attanasio et~al.}]{nozza2023really}
Debora Nozza, Giuseppe Attanasio, et~al. 2023.
\newblock {Is It Really That Simple? Prompting Language Models for Automatic
  Text Simplification in Italian}.
\newblock In \emph{CEUR Workshop Proceedings}. (seleziona).

\bibitem[{Ozdemir(2023)}]{ozdemir2023quick}
Sinan Ozdemir. 2023.
\newblock \emph{{Quick Start Guide to Large Language Models: Strategies and
  Best Practices for Using ChatGPT and Other LLMs}}.
\newblock Addison-Wesley Professional.

\bibitem[{Papineni et~al.(2002)Papineni, Roukos, Ward, and
  Zhu}]{papineni2002bleu}
Kishore Papineni, Salim Roukos, Todd Ward, and Wei-Jing Zhu. 2002.
\newblock {BLEU: A Method for Automatic Evaluation of Machine Translation}.
\newblock In \emph{Proceedings of the annual meeting of the Association for
  Computational Linguistics}, pages 311--318.

\bibitem[{Primpied et~al.(2022)Primpied, Beauchemin, and
  Khoury}]{Primpied2022Quantifying}
Vincent Primpied, David Beauchemin, and Richard Khoury. 2022.
\newblock Quantifying {French} {Document} {Complexity}.
\newblock \emph{Proceedings of the Canadian Conference on Artificial
  Intelligence}.
\newblock Https://caiac.pubpub.org/pub/iaeeogod.

\bibitem[{{Quebec}(2022{\natexlab{a}})}]{loicoderoute}
{Quebec}. 2022{\natexlab{a}}.
\newblock \href
  {https://www.legisquebec.gouv.qc.ca/fr/document/lc/C-24.2/20160908} {Code de
  la sécurité routière}.

\bibitem[{{Quebec}(2022{\natexlab{b}})}]{loiassauto}
{Quebec}. 2022{\natexlab{b}}.
\newblock \href {https://www.legisquebec.gouv.qc.ca/fr/document/lc/A-25} {Loi
  modifiant la loi sur l'assurance automobile, le code de la sécurité
  routière et d'autres dispositions}.

\bibitem[{Rebuffel et~al.(2021)Rebuffel, Scialom, Soulier, Piwowarski,
  Lamprier, Staiano, Scoutheeten, and Gallinari}]{rebuffel2021data}
Cl{\'e}ment Rebuffel, Thomas Scialom, Laure Soulier, Benjamin Piwowarski,
  Sylvain Lamprier, Jacopo Staiano, Geoffrey Scoutheeten, and Patrick
  Gallinari. 2021.
\newblock {Data-QuestEval: A Referenceless Metric for Data-To-Text Semantic
  Evaluation}.
\newblock In \emph{Conference on Empirical Methods in Natural Language
  Processing}, pages 8029--8036. Association for Computational Linguistics.

\bibitem[{Reimers and Gurevych(2019)}]{reimers2019sentencebert}
Nils Reimers and Iryna Gurevych. 2019.
\newblock {Sentence-BERT: Sentence Embeddings using Siamese BERT-Networks}.
\newblock In \emph{Proceedings of the Conference on Empirical Methods in
  Natural Language Processing and the International Joint Conference on Natural
  Language Processing}, pages 3982--3992.

\bibitem[{Ryan et~al.(2023)Ryan, Naous, and Xu}]{ryan2023revisiting}
Michael~J Ryan, Tarek Naous, and Wei Xu. 2023.
\newblock {Revisiting Non-english Text Simplification: A Unified Multilingual
  Benchmark}.
\newblock \emph{arXiv:2305.15678}.

\bibitem[{Saggion(2017)}]{saggion2017automatic}
Horacio Saggion. 2017.
\newblock {Automatic Text Simplification}.
\newblock \emph{Synthesis Lectures on Human Language Technologies},
  10(1):1--137.

\bibitem[{Shimorina and Belz(2021)}]{shimorina2021human}
Anastasia Shimorina and Anya Belz. 2021.
\newblock {The Human Evaluation Datasheet 1.0: A Template for Recording Details
  of Human Evaluation Experiments in NLP}.
\newblock \emph{arXiv:2103.09710}.

\bibitem[{{\v{S}}tajner(2021)}]{vstajner2021automatic}
Sanja {\v{S}}tajner. 2021.
\newblock {Automatic Text Simplification for Social Good: Progress and
  Challenges}.
\newblock \emph{Findings of the Association for Computational Linguistics:
  ACL-IJCNLP 2021}, pages 2637--2652.

\bibitem[{Stodden(2021)}]{stodden2021scale}
Regina Stodden. 2021.
\newblock {When the Scale is Unclear-Analysis of the Interpretation of Rating
  Scales in Human Evaluation of Text Simplification}.
\newblock In \emph{CTTS@ SEPLN}.

\bibitem[{Sulem et~al.(2018)Sulem, Abend, and Rappoport}]{sulem2018bleu}
Elior Sulem, Omri Abend, and Ari Rappoport. 2018.
\newblock {BLEU is Not Suitable for the Evaluation of Text Simplification}.
\newblock In \emph{Conference on Empirical Methods in Natural Language
  Processing}, pages 738--744. Association for Computational Linguistics.

\bibitem[{Tan et~al.(2023)Tan, Westermann, and Benyekhlef}]{tan2023chatgpt}
Jinzhe Tan, Hannes Westermann, and Karim Benyekhlef. 2023.
\newblock {ChatGPT as an Artificial Lawyer}.
\newblock \emph{Artificial Intelligence for Access to Justice}.

\bibitem[{Taylor(1953)}]{taylor1953cloze}
Wilson~L Taylor. 1953.
\newblock {“Cloze Procedure”: A New Tool for Measuring Readability}.
\newblock \emph{Journalism quarterly}, 30(4):415--433.

\bibitem[{T{\"o}rnberg(2024)}]{tornberg2024best}
Petter T{\"o}rnberg. 2024.
\newblock {Best Practices for Text Annotation with Large Language Models}.
\newblock \emph{arXiv:2402.05129}.

\bibitem[{van~der Lee et~al.(2019)van~der Lee, Gatt, van Miltenburg, Wubben,
  and Krahmer}]{van-der-lee-etal-2019-best}
Chris van~der Lee, Albert Gatt, Emiel van Miltenburg, Sander Wubben, and Emiel
  Krahmer. 2019.
\newblock \href {https://doi.org/10.18653/v1/W19-8643} {Best practices for the
  human evaluation of automatically generated text}.
\newblock In \emph{Proceedings of the International Conference on Natural
  Language Generation}, pages 355--368, Tokyo, Japan. Association for
  Computational Linguistics.

\bibitem[{Van~Hout and Vermeer(2007)}]{van2007comparing}
Roeland Van~Hout and Anne Vermeer. 2007.
\newblock {Comparing Measures of Lexical Richness}.
\newblock \emph{Modelling and assessing vocabulary knowledge}, 93:115.

\bibitem[{Weidinger et~al.(2021)Weidinger, Mellor, Rauh, Griffin, Uesato,
  Huang, Cheng, Glaese, Balle, Kasirzadeh et~al.}]{weidinger2021ethical}
Laura Weidinger, John Mellor, Maribeth Rauh, Conor Griffin, Jonathan Uesato,
  Po-Sen Huang, Myra Cheng, Mia Glaese, Borja Balle, Atoosa Kasirzadeh, et~al.
  2021.
\newblock {Ethical and Social Risks of Harm From Language Models}.
\newblock \emph{arXiv:2112.04359}.

\bibitem[{Wu and Arase(2024)}]{wu2024depth}
Xuanxin Wu and Yuki Arase. 2024.
\newblock {An In-depth Evaluation of GPT-4 in Sentence Simplification with
  Error-based Human Assessment}.
\newblock \emph{arXiv:2403.04963}.

\bibitem[{Xu et~al.(2015)Xu, Callison-Burch, and Napoles}]{xu2015problems}
Wei Xu, Chris Callison-Burch, and Courtney Napoles. 2015.
\newblock {Problems in Current Text Simplification Research: New Data Can
  Help}.
\newblock \emph{Transactions of the Association for Computational Linguistics},
  3:283--297.

\bibitem[{Zar(2005)}]{zar2005spearman}
Jerrold~H Zar. 2005.
\newblock {Spearman Rank Correlation}.
\newblock \emph{Encyclopedia of Biostatistics}, 7.

\bibitem[{Zhang et~al.(2019)Zhang, Kishore, Wu, Weinberger, and
  Artzi}]{zhang2019bertscore}
Tianyi Zhang, Varsha Kishore, Felix Wu, Kilian~Q Weinberger, and Yoav Artzi.
  2019.
\newblock {BERTScore: Evaluating Text Generation With BERT}.
\newblock In \emph{International Conference on Learning Representations}.

\end{thebibliography}
